\documentclass[journal,twoside]{IEEEtran}
\usepackage{cite}
\usepackage{amsmath,amssymb,amsfonts}
\usepackage{graphicx}
\usepackage{textcomp}
\usepackage{multirow}
\usepackage{booktabs}
\usepackage{tikz}
\usepackage{pgfplots}
\pgfplotsset{compat=1.18}

\definecolor{mvblue}{HTML}{0072B2}
\definecolor{smorange}{HTML}{D55E00}

\usepackage{hyperref}
\hypersetup{hidelinks}
\usepackage{orcidlink}
\usepackage{array}

\providecommand{\journalname}{Preprint}
\begin{document}

\title{Trading Parameters for Accuracy in MambaVision for Chest Radiograph Classification}

\author{Qaiser Farooq\,\orcidlink{0009-0009-9847-5285} and Tehseen Zia\,\orcidlink{0000-0001-8176-3373}\thanks{This work has been submitted to the IEEE for possible publication. Copyright may be transferred without notice, after which this version may no longer be accessible.

This work received no external funding. (Corresponding author: Tehseen Zia.)

Q.~Farooq and T.~Zia are with the Department of Computer Science, COMSATS University Islamabad, Islamabad, Pakistan (e-mail: qaiser.farooq@comsats.edu.pk; tehseen.zia@comsats.edu.pk).}}

\maketitle

\begin{abstract}
MambaVision, a hybrid state-space and attention backbone, places most of its parameters in its widest, fourth stage. We quantified what that stage contributes to multi-label chest radiograph classification. The fourth stage was replaced with lighter single-scan state-space blocks adapted from Spatial-Mamba, with multi-scale dilated depthwise convolutions applied to the scan output; the first three stages were retained. Tiny, Small, and Base models were evaluated across nine dataset--scale combinations drawn from two source cohorts: the National Institutes of Health ChestX-ray14 archive (full ChestMNIST and a high-resolution subset) and Indiana University Open-i. At the Tiny scale, replacement reduced parameter count by 35.7\% and increased graphics processing unit throughput by 19.3\%; the savings came from the stage's dense projections, not from removing quadratic attention. Without distillation, the mean area under the receiver operating characteristic curve on ChestMNIST was 0.004 lower than the baseline (95\% bootstrap confidence interval, 0.002 to 0.006), and 0.003 to 0.006 lower across all nine combinations. Knowledge distillation from an unreplaced teacher narrowed the gap to the undistilled baseline to 0.001--0.002 and met the equivalence criterion on ChestMNIST, but distilling the baseline with the same recipe improved it by 0.001--0.003, so a gap of 0.002--0.005 persisted under symmetric training. At the Base scale, the unreplaced Small model was both smaller and more accurate. Fourth-stage replacement is a useful trade-off only for Tiny and Small models. These measurements let developers of chest radiograph screening tools weigh model size and speed against a quantified loss of diagnostic discrimination.
\end{abstract}


\section{Introduction}
\label{sec:introduction}

\IEEEPARstart{A}{utomated} multi-label classification of frontal chest radiographs supports clinical triage, workflow prioritization, and population-scale screening \cite{ref42, ref44, ref46}. A thoracic classifier must resolve small, localized lesions such as pulmonary nodules or apical pneumothoraces while integrating long-range relationships across the lung fields, cardiac silhouette, and costophrenic angles. Convolutional neural networks (CNNs) such as ResNet \cite{ref27}, DenseNet \cite{ref24}, EfficientNet \cite{ref16}, and ConvNeXt \cite{ref22} capture local texture efficiently but aggregate context only within bounded neighborhoods, requiring deep stacking or large kernels \cite{ref14, ref15}. Vision transformers (ViTs) \cite{ref28, ref9, ref10} obtain global receptive fields through multi-head self-attention (MHSA) at $\mathcal{O}(N^{2}d)$ token-mixing cost and $\mathcal{O}(C^{2})$ parameter cost in channel width $C$ \cite{ref12}. Selective state space models (Mamba) \cite{ref4, ref3, ref7, ref8} offer linear-time alternatives that scan 2D token sequences with content-aware recurrence \cite{ref6, ref5}, and MambaVision \cite{ref1} combines residual convolutions in Stages~1--2 with interleaved MambaVision mixers and MHSA in Stages~3--4.

MambaVision is heavily imbalanced across stages. Because channel width doubles at each downsampling ($C \rightarrow 8C$), the dense projections of Stage~4 scale as $\mathcal{O}((8C)^{2})$. At $224\times 224$, Stage~4 processes only $7\times 7 = 49$ tokens, so its attention cost is negligible, yet its mixers, MHSA projections, and $4\times$ multilayer perceptrons (MLPs) hold $58.1\%$--$64.9\%$ of all parameters ($18.11\text{M}$ of $31.16\text{M}$ in MambaVision-T). Even at $512\times 512$ ($16\times 16 = 256$ Stage-4 tokens), dense projections rather than quadratic token mixing dominate Stage-4 compute, while Stage~3 ($1{,}024$ tokens) absorbs the primary attention cost.

Whether this parameter-heavy fourth stage is necessary for thoracic radiograph classification is a practical engineering question. Although a $31\text{M}$-parameter backbone is not prohibitive on datacenter hardware---both Tiny models exceed $1{,}000$ images/s on a single NVIDIA L40S at $224\times 224$---increasing throughput by $19.3\%$ and cutting checkpoint size by $35.7\%$ at a quantified $0.004$ Macro-AUC cost is a concrete trade-off for high-volume retrospective screening and shared hospital inference servers, and the throughput advantage widens to $21.8\%$--$23.4\%$ at $512\times 512$ resolution (Table~\ref{tab:highres}). Stage replacement also trims peak mixed-precision training memory by $6.6\%$--$11.7\%$ (e.g., from $11{,}950$ to $10{,}580$~MiB for Small at $512\times 512$ on a 12-GB consumer GPU, though halving the batch size with gradient accumulation can also accommodate the unreplaced baseline). We therefore characterize the exact efficiency--discrimination trade-off of replacing Stage~4 so that practitioners can decide when the savings justify the accuracy cost.

We test two falsifiable hypotheses: (H1) that the hybrid Stage~4 of MambaVision can be replaced by a lighter single-scan state-space stage with multi-scale dilated depthwise fusion adapted from \cite{ref2} without statistically significant loss of discrimination (within a $\pm 0.005$ equivalence margin in Macro-AUC); and (H2) that dilated spatial fusion (SASF) and selective scanning each contribute measurably over structured-pruning, parameter-matched reduced-MLP, or purely convolutional alternatives. Our contributions are:
\begin{itemize}
\item A stage-wise parameter, FLOP, and activation-memory characterization of MambaVision at $224\times 224$ and $512\times 512$ (Table~\ref{tab:stagewise}), demonstrating that Stage~4 holds $58.1\%$--$64.9\%$ of parameters, Stages~1--2 dominate peak inference activation memory, and dense projections dominate Stage-4 FLOPs at both resolutions.
\item A controlled architectural evaluation of Stage-4 replacement (\textbf{Spatial-MambaVision}), showing that substituting single-scan state-space blocks cuts total parameters by $19.7\%$--$35.7\%$ and Tiny Stage-4 FLOPs by $54.0\%$ ($0.892 \rightarrow 0.410$ GFLOPs) at a modest but consistent $0.003$--$0.006$ Macro-AUC cost, while delineating the scale boundary where replacement becomes Pareto-inferior to simply choosing a smaller unreplaced backbone.
\item A rigorous empirical analysis across nine dataset--scale cells drawn from two source cohorts (Table~\ref{tab:results}), paired with per-class clinical operating-point metrics (Table~\ref{tab:perclass}) and targeted ablations (Table~\ref{tab:ablation}) covering symmetric knowledge distillation, faithful hidden-state versus output-space state fusion, Taylor-based structured channel pruning, dilation schedules, and width-adaptation initialization.
\end{itemize}

\section{Related Work}
\label{sec:related_work}

\textit{Thoracic radiograph classification:} Large public databases including NIH ChestX-ray14 \cite{ref42}, Indiana Open-i \cite{ref43}, CheXpert \cite{ref46}, and MIMIC-CXR \cite{ref23} enabled deep thoracic classifiers. CheXNet \cite{ref44} popularized DenseNet-121 \cite{ref24} as a compact baseline ($6.97\text{M}$ parameters); subsequent work evaluated ResNet \cite{ref27}, ConvNeXt \cite{ref22, ref21}, and hierarchical transformers \cite{ref10, ref11, ref25}, while TorchXRayVision \cite{ref26} and MedMNIST v2/MedMNIST+ \cite{ref45, ref47} standardized evaluation protocols.

\textit{Efficient backbones, pruning, and distillation:} Lightweight CNNs such as MobileNetV3 \cite{ref17} and EfficientNet \cite{ref16} use inverted residuals, depthwise separable convolutions, and compound scaling; large-kernel ConvNets \cite{ref14, ref15} widen receptive fields without attention. Structured channel pruning \cite{ref51} reduces pretrained stages to a target budget without altering block topology, while knowledge distillation (KD) \cite{ref29, ref30} transfers teacher logits to compact students to recover capacity-related losses.

\textit{Visual state space models:} Vim \cite{ref6} introduced bidirectional scanning and VMamba \cite{ref5} four-way 2D selective scanning (SS2D). EfficientVMamba \cite{ref39}, LocalMamba \cite{ref38}, and GroupMamba \cite{ref41} reduce multi-scan latency through atrous, windowed, and channel-grouped scans. In medical imaging, MedMamba \cite{ref34} coupled grouped convolutions with SS2D, and 2DMamba \cite{ref33} adapted 2D scans to gigapixel histopathology. MambaVision \cite{ref1} showed that hybrid Mamba--Transformer deep stages outperform pure SSM backbones, and Spatial-Mamba \cite{ref2} showed that a single scan followed by structure-aware fusion of neighboring hidden states (SASF) matches or surpasses four-way scanning at lower cost. Our default Stage-4 block adopts the single-scan design but applies dilated fusion to the scan output for kernel fusion compatibility (Section~\ref{subsec:sassm}); we compare this output-space variant directly against the faithful hidden-state operator of \cite{ref2} in Table~\ref{tab:ablation}.

\begin{figure*}[!t]
\centering
\includegraphics[width=\textwidth]{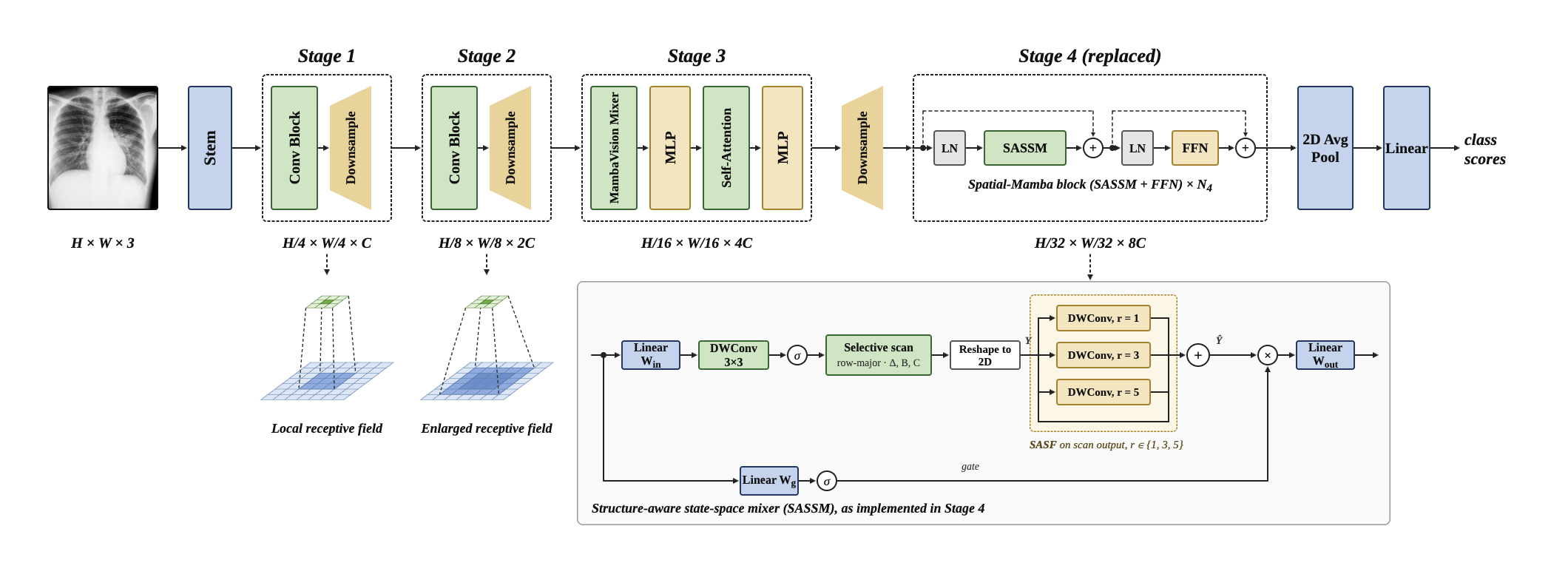}
\caption{Overview of the studied stage-replacement architecture (Spatial-MambaVision). The convolutional stem and residual convolutional Stages~1--2 extract local features; Stage~3 retains the MambaVision mixer and multi-head self-attention (MHSA) at width $4C$. Stage~4 replaces the original hybrid blocks with single-scan state-space blocks adapted from Spatial-Mamba. The detail box depicts the default SASSM variant, which applies dilated depthwise SASF convolutions to the reshaped scan output.}
\label{fig:pipeline}
\end{figure*}

\section{Method}
\label{sec:method}

\subsection{Preliminaries}
\label{subsec:preliminaries}

\textit{Selective state space models:} A linear SSM \cite{ref4} maps an input $u(t)$ to an output $y(t)$ through a latent state $\mathbf{h}(t) \in \mathbb{R}^{d_{\mathrm{state}}}$, $\mathbf{h}'(t) = \mathbf{A}\mathbf{h}(t) + \mathbf{B}u(t)$, $y(t) = \mathbf{C}\mathbf{h}(t)$. Discretizing with step $\Delta$ via zero-order hold (ZOH) gives
\begin{equation}
\overline{\mathbf{A}} = \exp(\Delta\mathbf{A}), \quad
\overline{\mathbf{B}} = (\Delta\mathbf{A})^{-1}\!\left(\exp(\Delta\mathbf{A})-\mathbf{I}\right)\Delta\mathbf{B},
\label{eq:ssm-disc}
\end{equation}
so that $\mathbf{h}_{t} = \overline{\mathbf{A}}\mathbf{h}_{t-1} + \overline{\mathbf{B}}u_{t}$ and $y_{t} = \mathbf{C}\mathbf{h}_{t}$. Mamba \cite{ref3} makes $\mathbf{B}_{t}$, $\mathbf{C}_{t}$, and $\Delta_{t}$ input-dependent and evaluates the recurrence across $D$ channels in $\mathcal{O}(NDd_{\mathrm{state}})$ time via a parallel scan.

\textit{MambaVision hybrid mixer:} In Stages~3--4, MambaVision \cite{ref1} interleaves MHSA \cite{ref9, ref28} with a dual-branch mixer:
\begin{equation}
\mathbf{Z} = \big[\mathrm{SSM}(\sigma(\mathrm{Conv}(\mathbf{X}\mathbf{W}_{1})))\;\Vert\;\sigma(\mathrm{Conv}(\mathbf{X}\mathbf{W}_{2}))\big]\,\mathbf{W}_{3},
\label{eq:mv}
\end{equation}
where $\sigma$ is SiLU, $\Vert$ is channel concatenation, and $\mathbf{W}_{1,2,3}$ are dense $\mathcal{O}(d^{2})$ projections followed by a $4\times$ MLP. MHSA costs $\mathcal{O}(N^{2}d)$ FLOPs, but its projections $\mathbf{W}_{Q,K,V,O}$ and MLP hold $\mathcal{O}(d^{2})$ parameters, which dominate storage when $d = 8C$.

\subsection{Overall Hierarchy and Stages 1--3}
\label{subsec:hierarchy}
As illustrated in Fig.~\ref{fig:pipeline}, a frontal radiograph $\mathbf{I}\in\mathbb{R}^{H\times W\times 3}$ (grayscale replicated to three channels) passes through a stem of two stride-2 $3\times 3$ convolutions to $\frac{H}{4}\times\frac{W}{4}\times C$. Four stages then halve resolution and double width, producing $\frac{H}{2^{i+1}}\times\frac{W}{2^{i+1}}\times 2^{i-1}C$ features ($i = 1,\dots,4$) with stride-2 convolution and batch normalization (BN) between stages. Stages~1--2 use residual $3\times 3$ Conv--BN--GELU blocks. Stage~3 ($\frac{H}{16}\times\frac{W}{16}$; $196$ tokens at $224^{2}$, $1{,}024$ at $512^{2}$; width $4C$) applies the mixer \eqref{eq:mv} in its first $\lceil N_{3}/2 \rceil$ blocks and MHSA in the remaining $\lfloor N_{3}/2 \rfloor$ ($29.0\%$ of MambaVision-T parameters). Stage~3 and baseline Stage~4 use unwindowed MHSA with windows equal to the stage grid: $14^{2}$ and $7^{2}$ at $224\times 224$, and $32^{2}$ and $16^{2}$ at $512\times 512$. Table~\ref{tab:stages} lists depths, widths, and Stage-4 settings; the complete stage-level specification is in the Supplement.

\begin{table}[!t]
\centering
\caption{Architecture Settings Across Scales and Stage-4 Parameter Footprint (Baseline MambaVision vs.\ Spatial-Mamba Replacement).}
\label{tab:stages}
\scriptsize
\setlength{\tabcolsep}{1.6pt}
\renewcommand{\arraystretch}{1.0}
\begin{tabular}{ccccccc}
\toprule
\textbf{Scale} & $C$/$8C$ & \textbf{Depths} & \textbf{FFN} & \textbf{SSM} & \textbf{SASF} & \textbf{S4 Params (MV/Ours)}\\
\midrule
Tiny & $80$ / $640$ & $1/3/8/4$ & $0.75\times$ & $0.875\times$ & $\{1,3,5\}$ & $18.11\text{M}$ / $7.00\text{M}$\\
Small & $96$ / $768$ & $3/3/7/5$ & $1.00\times$ & $1.000\times$ & $\{1,3,5\}$ & $32.04\text{M}$ / $14.80\text{M}$\\
Base & $128$ / $1024$ & $3/3/10/5$ & $1.375\times$ & $1.500\times$ & $\{1,3,5\}$ & $56.89\text{M}$ / $37.85\text{M}$\\
\bottomrule
\end{tabular}
\par\smallskip\scriptsize
\noindent Depths are blocks in Stages~1/2/3/4. Inner widths (FFN / SSM): Tiny $480 / 560$; Small $768 / 768$; Base $1408 / 1536$. $d_{\mathrm{state}} = 16$ at all scales.
\end{table}

\begin{table}[!t]
\centering
\caption{Stage-Wise Breakdown of MambaVision-T (Baseline) Versus Spatial-MambaVision-T (Ours). Live Activation Memory Measured Under \texttt{torch.no\_grad()} at $224\times 224$, Batch 64, FP16.}
\label{tab:stagewise}
\scriptsize
\setlength{\tabcolsep}{2.4pt}
\renewcommand{\arraystretch}{1.04}
\begin{tabular}{lcccc}
\toprule
& & \multicolumn{2}{c}{\textbf{GFLOPs}} & \textbf{Act.}\\
\cmidrule(lr){3-4}
\textbf{Stage / Module} & \textbf{Params (M)} & $224^{2}$ & $512^{2}$ & \textbf{(MiB)}\\
\midrule
Stem + downsamplers & $2.51$ ($8.1\%$) & $0.324$ & $1.693$ & $92$\\
Stage 1 (ResConv $\times 1$) & $0.12$ ($0.4\%$) & $0.362$ & $1.891$ & $122$\\
Stage 2 (ResConv $\times 3$) & $1.38$ ($4.4\%$) & $1.084$ & $5.663$ & $61$\\
Stage 3a (MV-Mixer $\times 4$) & $4.54$ ($14.6\%$) & $0.912$ & $4.764$ & $48$\\
Stage 3b (MHSA $\times 4$) & $4.50$ ($14.4\%$) & $0.926$ & $5.177$ & $44$\\
\midrule
Stage 4: Baseline ($\times 4$) & $18.11$ ($58.1\%$) & $0.892$ & $4.992$ & $24$\\
Stage 4: \textbf{Ours} ($\times 4$) & $\mathbf{7.00}$ ($\mathbf{34.9\%}$) & $\mathbf{0.410}$ & $\mathbf{2.142}$ & $\mathbf{17}$\\
\midrule
\textbf{Total: Baseline-T} & $31.16$ & $4.500$ & $24.180$ & $182^{\ast}$\\
\textbf{Total: Ours-T} & $\mathbf{20.05}$ ($-35.7\%$) & $\mathbf{4.018}$ & $\mathbf{21.330}$ & $\mathbf{161}^{\ast}$\\
\bottomrule
\end{tabular}
\par\smallskip\scriptsize
\noindent $^{\ast}$Peak live activation memory. Peak inference memory occurs in Stage~1 ($122$~MiB activations) plus resident FP16 weights ($59.4$ vs.\ $38.2$~MiB); full profiling details are in the Supplement.
\end{table}

\subsection{Stage-4 Replacement: Structure-Aware State-Space Mixer}
\label{subsec:sassm}
In Stage~4 (width $d = 8C$, grid $H_{4}\times W_{4}$), we replace all $N_{4}$ hybrid blocks with single-scan state-space blocks adapted from Spatial-Mamba \cite{ref2}. Each block applies a Structure-Aware State-Space Mixer (SASSM) and a feed-forward network (FFN) with pre-LayerNorm residuals:
\begin{equation}
\mathbf{Z} = \mathbf{X} + \mathrm{SASSM}\big(\mathrm{LN}(\mathbf{X})\big), \quad
\mathbf{X}_{\text{out}} = \mathbf{Z} + \mathrm{FFN}\big(\mathrm{LN}(\mathbf{Z})\big).
\label{eq:m-block}
\end{equation}
Let $\mathbf{U} = \mathrm{LN}(\mathbf{X})$ and let $D = \rho_{\mathrm{ssm}}d$ be the inner SSM width (Table~\ref{tab:stages}). Dense layers $\mathbf{W}_{\text{in}}, \mathbf{W}_{g} \in \mathbb{R}^{d \times D}$ produce a scan branch and a gate:
\begin{equation}
\mathbf{U}_{\text{scan}} = \sigma\big(\mathrm{DWConv}_{3\times 3}(\mathbf{U}\mathbf{W}_{\text{in}})\big), \qquad
\mathbf{G} = \sigma(\mathbf{U}\mathbf{W}_{g}).
\label{eq:sassm_proj}
\end{equation}
$\mathbf{U}_{\text{scan}}$ is flattened in row-major order into $\{\mathbf{u}_{t}\}_{t=1}^{L}$, $L = H_{4}W_{4}$, with input-dependent parameters $\mathbf{B}_{t} = \mathbf{W}_{B}\mathbf{u}_{t}$, $\mathbf{C}_{t} = \mathbf{W}_{C}\mathbf{u}_{t} \in \mathbb{R}^{d_{\mathrm{state}}}$ ($d_{\mathrm{state}} = 16$), and $\boldsymbol{\Delta}_{t} = \mathrm{softplus}(\mathbf{W}_{\Delta}\mathbf{u}_{t} + \mathbf{b}_{\Delta}) \in \mathbb{R}^{D}$. Discretizing the diagonal $\mathbf{A} \in \mathbb{R}^{D \times d_{\mathrm{state}}}$ and $\mathbf{B}_{t}$ via \eqref{eq:ssm-disc} gives $\overline{\mathbf{A}}_{t}, \overline{\mathbf{B}}_{t}$ and a single row-major scan over the state $\mathbf{h}_{t} \in \mathbb{R}^{D \times d_{\mathrm{state}}}$:
\begin{equation}
\begin{aligned}
\mathbf{h}_{t} &= \overline{\mathbf{A}}_{t} \odot \mathbf{h}_{t-1} + \overline{\mathbf{B}}_{t} \odot (\mathbf{u}_{t}\mathbf{1}^{\top}),\\
\mathbf{y}_{t} &= \mathbf{h}_{t}\mathbf{C}_{t} + \mathbf{d}_{\text{skip}} \odot \mathbf{u}_{t}.
\end{aligned}
\label{eq:sassm_scan}
\end{equation}
The fused CUDA kernel (\texttt{selective\_scan\_fn}) contracts $\mathbf{h}_{t}\mathbf{C}_{t}$ in on-chip SRAM without materializing $\mathbf{h}_{t}$, returning $\{\mathbf{y}_{t}\}$, which are reshaped to the 2D map $\mathbf{Y} \in \mathbb{R}^{H_{4}\times W_{4}\times D}$. Because row-major scanning separates vertically and diagonally adjacent lung regions, a structure-aware fusion module (SASF) of three parallel $3\times 3$ depthwise convolutions with dilations $r_{k} \in \{1, 3, 5\}$ is applied to $\mathbf{Y}$, followed by gating and output projection $\mathbf{W}_{\text{out}} \in \mathbb{R}^{D \times d}$:
\begin{equation}
\begin{aligned}
\hat{\mathbf{Y}} &= \mathbf{Y} + \textstyle\sum_{k=1}^{3}\mathrm{DWConv}_{3\times 3}^{(r_{k})}(\mathbf{Y}),\\
\mathrm{SASSM}(\mathbf{U}) &= \big(\hat{\mathbf{Y}} \odot \mathbf{G}\big)\mathbf{W}_{\text{out}}.
\end{aligned}
\label{eq:m-sasf}
\end{equation}
This output-space fusion departs from \cite{ref2}, which applies the dilated depthwise convolutions directly to the 4D latent states $\mathbf{h}_{t} \in \mathbb{R}^{H_{4}\times W_{4}\times D \times d_{\mathrm{state}}}$ prior to contracting with the input-dependent projection $\mathbf{C}_{t}$. Because $\mathbf{C}_{t}$ varies across spatial tokens, the two operators do not commute. To test whether this implementation choice sacrifices discrimination, we also implement the faithful hidden-state SASF operator using the unfused reference recurrence (\texttt{selective\_scan\_ref}) and benchmark it directly against \eqref{eq:m-sasf} in Table~\ref{tab:ablation} (Row~F).

\textit{Parameter cost, receptive field, and initialization:} Each depthwise branch adds $9D$ parameters ($5{,}040$ per block in Tiny), so the three-branch SASF module adds only $27D = 15{,}120$ parameters per block ($0.06\text{M}$ across all four Tiny Stage-4 blocks). Stage-4 savings therefore come from replacing the dual-branch mixers and $\mathbf{W}_{Q,K,V,O}$ projections ($5.00\text{M}$ combined in Tiny) and $4\times$ MLPs ($13.11\text{M}$) with narrower $\mathcal{O}((8C)\cdot D)$ projections and $0.75\times$--$1.375\times$ FFNs, cutting Tiny Stage-4 FLOPs by $54.0\%$ ($0.892 \rightarrow 0.410$ GFLOPs). Dilations $\{1,3,5\}$ give nominal $3\times 3$, $7\times 7$, and $11\times 11$ receptive fields; on the $7\times 7$ grid at $224\times 224$ the $r=5$ branch reaches the border from the center token and operates largely on padding, which we test by ablation. Stages~1--3 load official ImageNet-1K MambaVision weights \cite{ref1}, and Stage~4 loads compatible ImageNet-1K Spatial-Mamba weights \cite{ref2}. For Small ($8C=768$) and Base ($8C=1024$), outer Stage-4 widths match \texttt{Spatial-Mamba-S/B} directly; for Tiny, where \texttt{Spatial-Mamba-T} has Stage-4 width $512$ versus $640$ in MambaVision-T, we project the checkpoint weights via a ridge-regularized linear map $\mathbf{P} \in \mathbb{R}^{512\times 640}$ ($\lambda = 10^{-3}$) fit to pooled Stage-3/Stage-4 activations over $1{,}024$ ImageNet-1K images. Compared with zero-padding/truncation and random Stage-4 initialization, ridge mapping lowers feature reconstruction error ($0.064$ vs.\ $0.218$), improves 1-epoch linear-probe warmup Macro-AUC on ChestMNIST ($0.584$ vs.\ $0.538$ for zero-padding and $0.512$ for random Stage-4 initialization, where unaligned random Stage-4 weights scramble frozen Stage-3 representations), and accelerates Epoch-5 validation convergence ($0.771 \pm 0.004$ vs.\ $0.719 \pm 0.006$ for random init), but converges to a statistically indistinguishable final test AUC after $80$ epochs ($0.811 \pm 0.002$ vs.\ $0.809 \pm 0.003$, $p = 0.096$; Table~\ref{tab:ablation}, Row~8; full initialization table and formulation in the Supplement).

\subsection{Classification Head, Loss, and Knowledge Distillation}
\label{subsec:head}
The Stage-4 output is pooled by global average pooling and mapped to $K$ logits $\hat{\mathbf{y}}$. We minimize class-weighted binary cross-entropy (\texttt{BCEWithLogitsLoss} with \texttt{pos\_weight}) where $w_{c}^{+} = \min(N_{c}^{-}/N_{c}^{+}, w_{\max})$ scales only the positive term:
\begin{equation}
\mathcal{L}_{\text{BCE}} = -\frac{1}{K}\sum_{c=1}^{K}
\Big[w_{c}^{+}\,y_{c}\log\hat{p}_{c} + (1-y_{c})\log(1-\hat{p}_{c})\Big],
\label{eq:m-loss}
\end{equation}
with $\hat{p}_{c}=\sigma(\hat{y}_{c})$. To test whether knowledge distillation (KD) \cite{ref29, ref30} can recover the accuracy cost of stage replacement---and whether the unreplaced baseline benefits equally from the same supervision---we evaluate response-based KD using a fine-tuned MambaVision-B teacher ($\hat{\mathbf{y}}^{T}$) across all scales with binary Kullback--Leibler divergence at temperature $\tau = 2.0$ and balance weight $\alpha = 0.5$:
\begin{equation}
\mathcal{L}_{\text{total}} = (1-\alpha)\mathcal{L}_{\text{BCE}} + \alpha\tau^{2}\,\mathrm{KL}\big(\sigma(\hat{\mathbf{y}}^{T}/\tau)\,\Vert\,\sigma(\hat{\mathbf{y}}/\tau)\big).
\label{eq:kd}
\end{equation}
Using the common MambaVision-B teacher allows us to apply the exact same distillation recipe symmetrically to both Spatial-MambaVision-T and the baseline MambaVision-T (Tables~\ref{tab:results}, \ref{tab:sota}, and \ref{tab:ablation}). Distilling instead from a same-scale MambaVision-T teacher gives results within $0.001$ AUC of those obtained with the MambaVision-B teacher (Supplement).

\section{Experiments}
\label{sec:experiments}

\subsection{Datasets and Partitions}
We evaluate across nine dataset--scale cells drawn from two source cohorts---NIH ChestX-ray14 \cite{ref42} (at full population scale in ChestMNIST and in a high-resolution subset) and Indiana University Open-i \cite{ref43} (Table~\ref{tab:datasets}). Because both ChestMNIST and the NIH subset derive from the NIH ChestX-ray14 archive, they share imaging equipment, patient demographics, and NLP label-extraction noise; thus, the nine dataset--scale evaluations represent two independent clinical source cohorts rather than three. All data are publicly available, de-identified research datasets; no new human-subject data were collected for this study.

\textit{ChestMNIST:} The official $224\times 224$ MedMNIST+ release (\texttt{chestmnist\_224}) \cite{ref45, ref47} of all $112{,}120$ NIH radiographs with $14$ labels and the standard patient-disjoint split ($16{,}510$ positive test labels).

\textit{Indiana Open-i:} $3{,}818$ frontal radiographs with XML reports. Negation-aware NLP extraction and exclusion of three classes with $<30$ positives (Fibrosis, Pleural Thickening, Hernia) leave $11$ pathologies ($2{,}497$ positive labels; $55.5\%$ normal studies). Because the public Open-i XML release strips patient identifiers, splits are defined by report (\texttt{uid}). Cross-referencing exact patient age, sex, and comparison phrases (``compared with prior study...'') across the $764$ test reports identifies $68$ studies ($8.9\%$) that reference a prior examination, bounding potential cross-split patient overlap at $\le 8.9\%$ of test studies. An initial unstratified split (seed $42$, $70/10/20$) gave unbalanced validation/test densities ($0.867$ / $0.503$), so we re-split the \texttt{uid}s with iterative multi-label stratification \cite{ref52} ($1{,}748$ / $250$ / $499$ positive labels) and report the stratified split; legacy-split results are in the Supplement.

\textit{NIH low-data subset:} To evaluate sample efficiency and $512\times 512$ scaling on native $1024\times 1024$ radiographs, we use a patient-disjoint stratified sample of $7{,}121$ NIH radiographs over $13$ classes (Hernia excluded; $2{,}655$ positive test labels). All models on this benchmark are initialized strictly from ImageNet-1K weights and trained only on its $3{,}176$ training images, so there is zero within-experiment leakage; however, because it is sampled from the same NIH archive, $2{,}514$ of its $3{,}592$ test radiographs ($70.0\%$) appear in the official ChestMNIST training split. Capping training at $3{,}176$ images ($44.6\%$; $353$ validation, a $9:1$ ratio) simulates a resource-constrained hospital cohort, while allocating $3{,}592$ images ($50.4\%$) to evaluation keeps rare classes such as Fibrosis ($n^{+} = 58$) and Edema ($n^{+} = 68$) large enough for narrow bootstrap intervals. Pneumonia ($n^{+} = 43$) remains below our $n^{+} \ge 50$ threshold and is excluded from per-class claims.

\begin{table}[!t]
\centering
\caption{Datasets, Partitions, Positive-Label Density (Labels per Radiograph), and Test Count $n^{+}$ of the Rarest Class.}
\label{tab:datasets}
\scriptsize
\setlength{\tabcolsep}{1.3pt}
\renewcommand{\arraystretch}{1.0}
\begin{tabular}{lcccc}
\toprule
\textbf{Dataset} & $K$ & \textbf{Train / Val / Test} & \textbf{Density (Tr/Va/Te)} & \textbf{Rarest $n^{+}$}\\
\midrule
ChestMNIST & $14$ & $78{,}468$ / $11{,}219$ / $22{,}433$ & $0.708$ / $0.705$ / $0.736$ & $42$ (Hernia)\\
Indiana (strat.) & $11$ & $2{,}672$ / $382$ / $764$ & $0.654$ / $0.654$ / $0.653$ & $14$ (Mass)\\
NIH subset & $13$ & $3{,}176$ / $353$ / $3{,}592$ & $0.741$ / $0.739$ / $0.739$ & $43$ (Pneum.)\\
\bottomrule
\end{tabular}
\par\smallskip\scriptsize
\noindent ChestMNIST and the NIH subset are drawn from the same NIH ChestX-ray14 cohort and are strictly patient-disjoint within each benchmark. Indiana is split by report \texttt{uid} ($\le 8.9\%$ potential repeat-visit overlap); the legacy unstratified Indiana split is reported in the Supplement.
\end{table}

\subsection{Training, Measurement, and Statistics}
All models are trained for $80$ epochs with AdamW \cite{ref49} (5-epoch warmup, cosine decay, FP16 AMP), batch $64$ ($32$ at $512\times 512$), backbone learning rate $1\times 10^{-4}$ ($1.5\times 10^{-4}$ on Indiana), head learning rate $5\times 10^{-4}$, weight decay $5\times 10^{-2}$ ($1\times 10^{-2}$ on NIH), random resized crop ($0.85$--$1.0$), and horizontal flip ($p=0.5$); $w_{\max}$ is $15$ on ChestMNIST/NIH and $10$ on Indiana. Checkpoints are selected by validation Macro-AUC. All baselines in Table~\ref{tab:sota} received the same hyperparameter search budget over learning rate and weight decay (grid in the Supplement).

\textit{Efficiency measurement:} Models are profiled in PyTorch 2.4 without the classification head. GFLOPs (batch $1$) and parameters are measured with \texttt{fvcore}, using custom handles for the selective scan and SASF convolutions. Training memory (batch $64$, AMP) and FP16 inference throughput (mean $\pm$ SD over $5$ runs of $200$ synchronized iterations after $50$ warm-up steps, batch $64$) are measured on an NVIDIA L40S (48~GB) and an RTX 3060 (12~GB); CPU latency (batch $1$, FP32, 4 threads) uses \texttt{selective\_scan\_ref} on an Intel Xeon Gold 6338 as a reference only. An 80-epoch ChestMNIST run on one L40S takes $4.6$ / $7.0$ / $12.5$~h (Tiny/Small/Base) for MambaVision and $3.9$ / $6.0$ / $11.2$~h for Spatial-MambaVision; Indiana and NIH-subset runs take $0.18$--$0.55$~h at $224\times 224$ and $0.85$--$1.45$~h at $512\times 512$.

\textit{Metrics and operating points:} We lead with threshold-independent Macro-AUC and mean average precision (mAP) because positive-label prevalence in multi-label thoracic screening is low ($5.2\%$ average across ChestMNIST classes, ranging from $0.19\%$ for Hernia to $17.6\%$ for Infiltration) and Eq.~\eqref{eq:m-loss} upweights positive instances up to $w_{\max}=15$. Under class-weighted training, an uncalibrated fixed threshold of $\hat{p}_{c} \ge 0.5$ produces high recall but low precision (Macro-F1 $\approx 0.24$; full thresholded metrics in the Supplement). Because clinical deployment calibrates per-pathology thresholds to a target false-positive rate, Table~\ref{tab:perclass} also reports per-class sensitivity at a fixed $90\%$ specificity ($\mathrm{Sens@90\%Spec}$) alongside absolute per-class AUC on ChestMNIST.

\textit{Statistical inference:} Each configuration uses $10$ random seeds (except Row~F in Table~\ref{tab:ablation}, which uses $3$ seeds due to unfused recurrence cost). Because Stages~1--3 (and pretrained Stage-4 weights, except Row~8) always initialize from the same fixed ImageNet-1K checkpoint, seeds vary only minibatch ordering, augmentation realizations, and linear-head initialization; the reported across-seed SD therefore understates full pretraining-plus-finetuning variance. We report two complementary inferential quantities that address distinct targets: (i) a two-level bootstrap 95\% confidence interval ($2{,}000$ resamples over both seeds and test radiographs), which estimates generalization uncertainty over new patient cohorts drawn from the same population, and (ii) Welch's two-sample $t$-test across seed means, which tests whether two training procedures converge to different expected values on the fixed test set. We lead comparative claims with the bootstrap 95\% CIs. Equivalence is assessed via two one-sided tests (TOST) \cite{ref53} at a $\pm 0.005$ Macro-AUC margin for undistilled ($p_{\text{TOST}}$) and distilled ($p_{\text{TOST}}^{\text{KD}}$) models. Across the nine dataset--scale cells of Table~\ref{tab:results}, Holm--Bonferroni multiplicity adjustment \cite{ref54} is applied to the two-sided difference tests ($p_{\text{Holm}}$), whereas $p_{\text{TOST}}$ and $p_{\text{TOST}}^{\text{KD}}$ are reported unadjusted per cell (all three ChestMNIST $p_{\text{TOST}}^{\text{KD}}$ values also remain $\le 0.042$ if Holm-adjusted across the nine cells). Per-class DeLong tests \cite{ref55} and prediction logs are provided in the Supplement and public repository.

\begin{table*}[!t]
\centering
\caption{Computational Complexity and Hardware Efficiency at $224\times224$ Input. GFLOPs via \texttt{fvcore} (Batch 1); Peak AMP Training Memory (Batch 64); GPU Throughput ($\text{Mean} \pm \text{SD}$ Over 5 Launches, Batch 64, FP16) on an NVIDIA L40S and Consumer RTX 3060; CPU Latency at Batch 1 (FP32, 4 Threads).}
\label{tab:complexity}
\scriptsize
\setlength{\tabcolsep}{3.5pt}
\renewcommand{\arraystretch}{1.0}
\begin{tabular}{llcccccc}
\toprule
\textbf{Scale} & \textbf{Backbone} & \textbf{Params (M)} $\downarrow$ & \textbf{GFLOPs} $\downarrow$ & \textbf{Train Mem.\ (MiB)} $\downarrow$ & \textbf{L40S (img/s)} $\uparrow$ & \textbf{RTX 3060 (img/s)} $\uparrow$ & \textbf{CPU$^{\ast}$ (ms/img)} $\downarrow$\\
\midrule
\multirow{3}{*}{Tiny}
 & MambaVision-T \cite{ref1} & $31.16$ & $4.500$ & $3{,}840$ & $1{,}184 \pm 15$ & $346 \pm 5$ & $28.4$\\
 & \textbf{Spatial-MambaVision-T (Ours)} & $\mathbf{20.05}$ & $\mathbf{4.018}$ & $\mathbf{3{,}490}$ & $\mathbf{1{,}412 \pm 18}$ & $\mathbf{411 \pm 6}$ & $\mathbf{23.1}$\\
 & \textit{Change} & $-35.7\%$ & $-10.7\%$ & $-9.1\%$ & $+19.3\%$ & $+18.8\%$ & $-18.7\%$\\
\midrule
\multirow{3}{*}{Small}
 & MambaVision-S \cite{ref1} & $49.38$ & $7.495$ & $5{,}920$ & $762 \pm 10$ & $221 \pm 3$ & $44.8$\\
 & \textbf{Spatial-MambaVision-S (Ours)} & $\mathbf{32.14}$ & $\mathbf{6.742}$ & $\mathbf{5{,}410}$ & $\mathbf{896 \pm 12}$ & $\mathbf{258 \pm 4}$ & $\mathbf{37.2}$\\
 & \textit{Change} & $-34.9\%$ & $-10.0\%$ & $-8.6\%$ & $+17.6\%$ & $+16.7\%$ & $-17.0\%$\\
\midrule
\multirow{3}{*}{Base}
 & MambaVision-B \cite{ref1} & $96.67$ & $14.975$ & $10{,}450$ & $418 \pm 6$ & $121 \pm 2$ & $86.5$\\
 & \textbf{Spatial-MambaVision-B (Ours)} & $\mathbf{77.63}$ & $\mathbf{14.198}$ & $\mathbf{9{,}760}$ & $\mathbf{465 \pm 7}$ & $\mathbf{134 \pm 2}$ & $\mathbf{76.9}$\\
 & \textit{Change} & $-19.7\%$ & $-5.2\%$ & $-6.6\%$ & $+11.2\%$ & $+10.7\%$ & $-11.1\%$\\
\bottomrule
\end{tabular}
\par\smallskip\scriptsize
\noindent $^{\ast}$CPU timings use the unoptimized TorchScript \texttt{selective\_scan\_ref} recurrence and are not representative of production deployment. FP32 checkpoint storage drops from $118.87\to76.49$~MiB (T), $188.37\to122.61$~MiB (S), and $368.78\to296.13$~MiB (B).
\end{table*}

\subsection{Computational Complexity and Memory}
Table~\ref{tab:complexity} summarizes the efficiency changes (bar charts are in the Supplement). \textit{Parameters vs.\ FLOPs:} parameters drop by $19.7\%$--$35.7\%$ but GFLOPs by only $5.2\%$--$10.7\%$, because Stage~4 holds $58.1\%$ of MambaVision-T parameters but $19.8\%$ of its GFLOPs ($0.892$ of $4.500$; Table~\ref{tab:stagewise}); within Stage~4, Tiny parameters fall by $61.3\%$ ($18.11\text{M} \to 7.00\text{M}$) and FLOPs by $54.0\%$. \textit{Inference vs.\ training memory:} peak live activation memory at $224\times 224$ occurs in Stage~1 ($122$~MiB in Tiny) and is freed before Stage~4 ($24$ vs.\ $17$~MiB), so the peak inference reduction ($21.2$ / $32.9$ / $36.3$~MiB for Tiny/Small/Base) reflects only resident FP16 weights. During AMP training, activations of all stages coexist with FP32 master weights, gradients, and AdamW states ($18$ bytes/parameter), yielding a $350$--$690$~MiB ($6.6\%$--$9.1\%$) reduction in peak training memory. \textit{Throughput:} FP16 throughput rises by $18.8\%$--$19.3\%$ (Tiny), $16.7\%$--$17.6\%$ (Small), and $10.7\%$--$11.2\%$ (Base) on both GPUs.

\begin{figure}[!t]
\centering
\begin{tikzpicture}
\begin{axis}[
    width=0.96\columnwidth,
    height=4.8cm,
    xlabel={Parameters (Millions) $\downarrow$},
    ylabel={ChestMNIST Macro-AUC $\uparrow$},
    xmin=3, xmax=36.5,
    ymin=0.783, ymax=0.8212,
    ytick={0.79,0.80,0.81,0.82},
    yticklabel style={/pgf/number format/fixed, /pgf/number format/precision=2, /pgf/number format/fixed zerofill},
    grid=both,
    grid style={dashed, gray!30},
    tick label style={font=\scriptsize},
    label style={font=\footnotesize\bfseries},
    legend columns=2,
    legend cell align=left, legend style={font=\scriptsize, draw=gray!50, at={(0.5,-0.31)}, anchor=north, /tikz/every even column/.append style={column sep=6pt}}
]
\addplot[only marks, mark=square*, mark size=2.3pt, color=gray!70!black, error bars/.cd, y dir=both, y explicit] coordinates {
    (4.22, 0.789) +- (0,0.003)
    (6.97, 0.793) +- (0,0.002)
    (7.70, 0.799) +- (0,0.002)
    (23.54, 0.790) +- (0,0.002)
    (27.83, 0.802) +- (0,0.002)
};
\addlegendentry{CNN Backbones}

\addplot[only marks, mark=triangle*, mark size=2.6pt, color=orange!85!black, error bars/.cd, y dir=both, y explicit] coordinates {
    (27.53, 0.800) +- (0,0.002)
    (14.50, 0.798) +- (0,0.003)
    (11.02, 0.801) +- (0,0.002)
    (25.80, 0.804) +- (0,0.002)
    (30.22, 0.806) +- (0,0.002)
};
\addlegendentry{ViT / Pure SSMs}

\addplot[only marks, mark=pentagon*, mark size=3.0pt, color=red!80!black, error bars/.cd, y dir=both, y explicit] coordinates {
    (31.16, 0.815) +- (0,0.002)
    (31.16, 0.817) +- (0,0.002)
};
\addlegendentry{MambaVision-T (plain / +KD)}

\addplot[only marks, mark=star, mark size=3.8pt, thick, color=teal!85!black, error bars/.cd, y dir=both, y explicit] coordinates {
    (20.05, 0.811) +- (0,0.002)
    (20.05, 0.814) +- (0,0.002)
};
\addlegendentry{Spatial-MambaVision-T (plain / +KD)}

\addplot[dashed, thick, color=teal!75!black, forget plot] coordinates {
    (4.22, 0.789) (6.97, 0.793) (7.70, 0.799) (11.02, 0.801) (20.05, 0.814) (31.16, 0.817)
};

\node[font=\tiny, anchor=north west] at (axis cs:4.6,0.7886) {MobileNetV3-L};
\node[font=\tiny, anchor=west] at (axis cs:7.5,0.7924) {DenseNet-121};
\node[font=\tiny, anchor=north west] at (axis cs:8.2,0.7988) {EffNet-B2};
\node[font=\tiny, anchor=south east] at (axis cs:10.6,0.8026) {EffVMamba-S};
\node[font=\tiny, anchor=north west] at (axis cs:14.9,0.7972) {MedMamba-T};
\node[font=\tiny, anchor=north] at (axis cs:23.54,0.7876) {ResNet-50};
\node[font=\tiny, anchor=north] at (axis cs:27.53,0.7977) {Swin-T};
\node[font=\tiny, anchor=north west] at (axis cs:28.3,0.8030) {ConvNeXt-T};
\node[font=\tiny, anchor=north east] at (axis cs:25.4,0.8034) {Spatial-Mamba-T};
\node[font=\tiny, anchor=west] at (axis cs:30.8,0.8062) {VMamba-T};
\node[font=\tiny\bfseries, text=teal!85!black, anchor=north west] at (axis cs:20.6,0.8108) {Ours-T (0.811)};
\node[font=\tiny\bfseries, text=teal!85!black, anchor=south east] at (axis cs:19.6,0.8144) {Ours-T+KD (0.814)};
\node[font=\tiny\bfseries, text=red!80!black, anchor=north east] at (axis cs:30.6,0.8143) {MV-T (0.815)};
\node[font=\tiny\bfseries, text=red!80!black, anchor=east] at (axis cs:30.3,0.8188) {MV-T+KD (0.817)};
\end{axis}
\end{tikzpicture}
\caption{Accuracy versus parameters (mean $\pm$ SD, 10 seeds) on ChestMNIST ($224\times 224$) for ImageNet-pretrained backbones; the dashed line marks the Pareto frontier with KD models included. Within each training regime (undistilled or with KD), Spatial-MambaVision-T ($20.05\text{M}$) lies on the frontier between EfficientVMamba-S and MambaVision-T. Domain-pretrained baselines are omitted because they were pretrained on the same source cohorts as our evaluation data (Section~\ref{subsec:results}).}
\label{fig:pareto}
\end{figure}
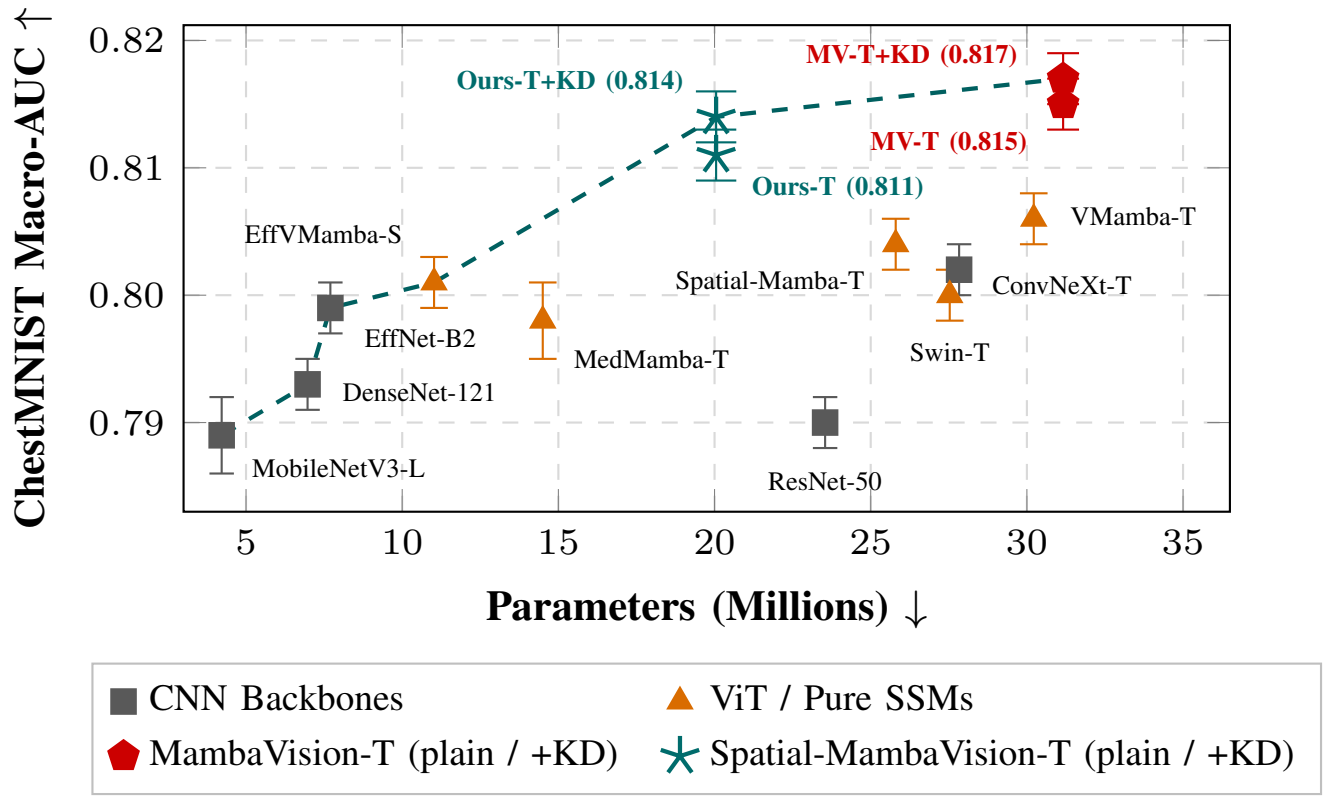

\begin{table*}[!t]
\centering
\caption{Full $3\times 3$ Test-Set Classification Matrix ($\text{Mean} \pm \text{SD}$, 10 Seeds) at $224\times 224$ Across Nine Dataset--Scale Cells Drawn From Two Source Cohorts. $\Delta$AUC Is Undistilled Baseline Minus Undistilled Ours With Two-Level Bootstrap 95\% CI; $p_{\text{Welch}}$ and Holm-Adjusted $p_{\text{Holm}}$ Test for Difference; Unadjusted $p_{\text{TOST}}$ and $p_{\text{TOST}}^{\text{KD}}$ Test Equivalence Within $\pm 0.005$ AUC Against the Undistilled Baseline.}
\label{tab:results}
\scriptsize
\setlength{\tabcolsep}{1.9pt}
\renewcommand{\arraystretch}{1.0}
\begin{tabular}{llccccccccc}
\toprule
\textbf{Dataset} & \textbf{Scale} & \textbf{MV AUC} & \textbf{Ours AUC} & \textbf{mAP (MV / Ours)} & \textbf{$\Delta$AUC [95\% CI]} & $p_{\text{Welch}}$ & $p_{\text{Holm}}$ & $p_{\text{TOST}}$ & \textbf{Ours+KD $\Delta$AUC [CI]} & $p_{\text{TOST}}^{\text{KD}}$\\
\midrule
\multirow{3}{*}{ChestMNIST}
 & Tiny & $0.815 \pm 0.002$ & $0.811 \pm 0.002$ & $0.226$ / $0.222$ & $0.004\;[0.002, 0.006]$ & $0.001$ & $0.008$ & $0.203$ & $0.001\;[-0.001, 0.003]$ & $0.001$\\
 & Small & $0.820 \pm 0.002$ & $0.817 \pm 0.002$ & $0.233$ / $0.229$ & $0.003\;[0.001, 0.005]$ & $0.006$ & $0.036$ & $0.047$ & $0.001\;[-0.001, 0.003]$ & $0.001$\\
 & Base & $0.823 \pm 0.002$ & $0.818 \pm 0.003$ & $0.238$ / $0.232$ & $0.005\;[0.003, 0.007]$ & $<0.001$ & $<0.001$ & $0.500$ & $0.002\;[0.000, 0.004]$ & $0.006$\\
\midrule
\multirow{3}{*}{Indiana}
 & Tiny & $0.754 \pm 0.007$ & $0.750 \pm 0.008$ & $0.241$ / $0.236$ & $0.004\;[-0.003, 0.011]$ & $0.253$ & $0.330$ & $0.388$ & $0.002\;[-0.005, 0.009]$ & $0.195$\\
 & Small & $0.761 \pm 0.007$ & $0.756 \pm 0.008$ & $0.249$ / $0.244$ & $0.005\;[-0.002, 0.012]$ & $0.165$ & $0.330$ & $0.500$ & $0.001\;[-0.006, 0.008]$ & $0.127$\\
 & Base & $0.764 \pm 0.008$ & $0.758 \pm 0.008$ & $0.253$ / $0.246$ & $0.006\;[-0.001, 0.013]$ & $0.105$ & $0.315$ & $0.606$ & $0.002\;[-0.005, 0.009]$ & $0.203$\\
\midrule
\multirow{3}{*}{NIH subset}
 & Tiny & $0.799 \pm 0.004$ & $0.795 \pm 0.004$ & $0.231$ / $0.227$ & $0.004\;[0.001, 0.008]$ & $0.026$ & $0.104$ & $0.289$ & $0.001\;[-0.002, 0.004]$ & $0.009$\\
 & Small & $0.805 \pm 0.004$ & $0.800 \pm 0.005$ & $0.238$ / $0.232$ & $0.005\;[0.001, 0.009]$ & $0.012$ & $0.060$ & $0.500$ & $0.002\;[-0.002, 0.006]$ & $0.066$\\
 & Base & $0.809 \pm 0.004$ & $0.803 \pm 0.005$ & $0.243$ / $0.237$ & $0.006\;[0.002, 0.010]$ & $0.004$ & $0.028$ & $0.683$ & $0.002\;[-0.002, 0.006]$ & $0.075$\\
\midrule
\multicolumn{2}{l}{\textbf{Symmetric KD control (Tiny):}} & \multicolumn{9}{l}{\textbf{MambaVision-T + KD} reaches $0.817 \pm 0.002$ (ChestMNIST), $0.757 \pm 0.007$ (Indiana), and $0.800 \pm 0.004$ (NIH subset);}\\
\multicolumn{2}{l}{\textit{MV-T+KD minus Ours-T+KD}} & \multicolumn{9}{l}{ChestMNIST: $0.003\;[0.001, 0.005]$ ($p_{\text{Welch}}=0.004$, $p_{\text{TOST}}=0.047$); Indiana: $0.005\;[-0.002, 0.012]$; NIH: $0.002\;[-0.002, 0.006]$.}\\
\bottomrule
\end{tabular}
\end{table*}

\subsection{Classification Across the \texorpdfstring{$3\times 3$}{3x3} Matrix and Baselines}
\label{subsec:results}
\begin{figure}[!t]
\centering
\begin{tikzpicture}
\begin{axis}[
    width=0.97\columnwidth,
    height=4.5cm,
    xlabel={$\Delta\text{AUC}$ (MambaVision $-$ Spatial-MambaVision)},
    xmin=-0.008, xmax=0.0145,
    scaled x ticks=false,
    xtick={-0.005,0,0.005,0.010},
    xticklabel style={/pgf/number format/fixed, /pgf/number format/precision=3, /pgf/number format/fixed zerofill},
    ymin=0.4, ymax=9.6,
    ytick={1,...,9},
    yticklabels={NIH-B, NIH-S, NIH-T, Indiana-B, Indiana-S, Indiana-T, ChestMNIST-B, ChestMNIST-S, ChestMNIST-T},
    tick label style={font=\scriptsize},
    label style={font=\footnotesize\bfseries},
    ymajorgrids, grid style={dashed, gray!25},
    axis lines*=left
]
\path[fill=gray!18] (axis cs:-0.005,0.4) rectangle (axis cs:0.005,9.6);
\draw[thick, dashed, black!70] (axis cs:0,0.4) -- (axis cs:0,9.6);
\addplot[only marks, mark=*, mark size=1.9pt, color=smorange,
    error bars/.cd, x dir=both, x explicit] coordinates {
    (0.004, 9.17) +- (0.002, 0)
    (0.003, 8.17) +- (0.002, 0)
    (0.005, 7.17) +- (0.002, 0)
    (0.004, 6.17) +- (0.007, 0)
    (0.005, 5.17) +- (0.007, 0)
    (0.006, 4.17) +- (0.007, 0)
    (0.004, 3.17) += (0.004, 0) -= (0.003, 0)
    (0.005, 2.17) +- (0.004, 0)
    (0.006, 1.17) +- (0.004, 0)
};
\addplot[only marks, mark=square*, mark size=1.7pt, color=mvblue,
    error bars/.cd, x dir=both, x explicit] coordinates {
    (0.001, 8.83) +- (0.002, 0)
    (0.001, 7.83) +- (0.002, 0)
    (0.002, 6.83) +- (0.002, 0)
    (0.002, 5.83) +- (0.007, 0)
    (0.001, 4.83) +- (0.007, 0)
    (0.002, 3.83) +- (0.007, 0)
    (0.001, 2.83) +- (0.003, 0)
    (0.002, 1.83) +- (0.004, 0)
    (0.002, 0.83) +- (0.004, 0)
};
\end{axis}
\end{tikzpicture}
\caption{Accuracy cost of Stage-4 replacement in the nine dataset--scale cells of Table~\ref{tab:results}: $\Delta\mathrm{AUC}$ of the undistilled same-scale MambaVision baseline minus Spatial-MambaVision without KD (circles) and with KD (squares), with two-level bootstrap 95\% CIs. The shaded band marks the $\pm 0.005$ equivalence margin. The comparison in which the baseline is also distilled is given in the bottom rows of Table~\ref{tab:results}.}
\label{fig:forest}
\end{figure}
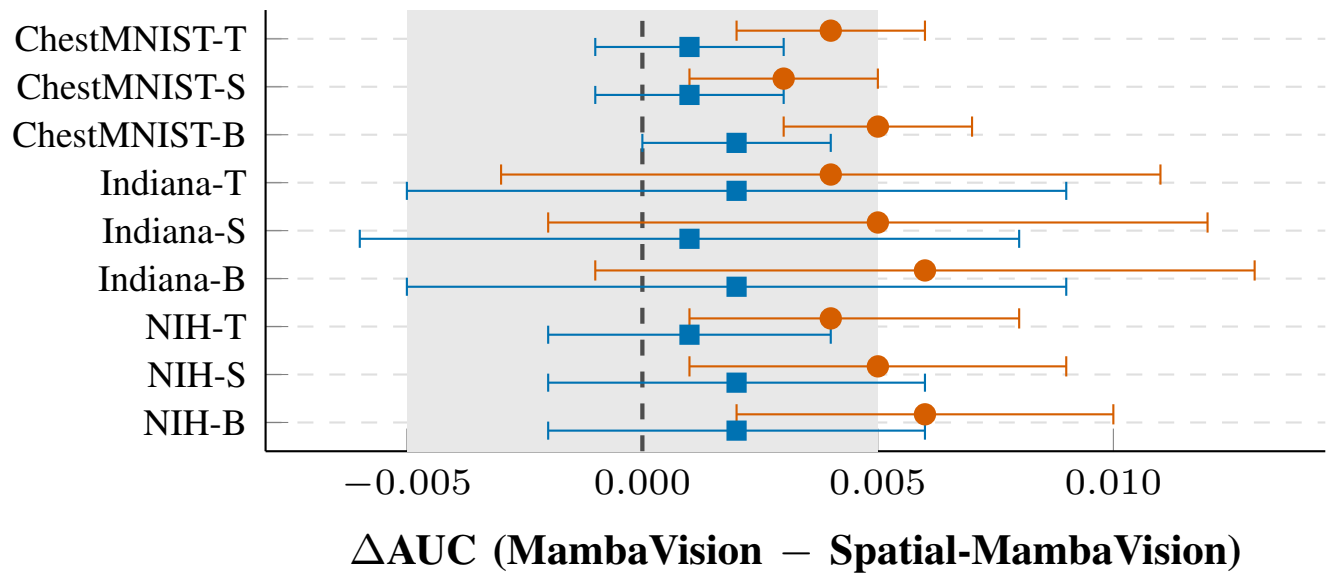

Across all nine dataset--scale cells of Table~\ref{tab:results} (summarized in Fig.~\ref{fig:forest}), undistilled Spatial-MambaVision trails its same-scale baseline by $\Delta\mathrm{AUC} \in [0.003, 0.006]$. On ChestMNIST ($N_{\text{test}} = 22{,}433$), the bootstrap 95\% CIs exclude zero at all three scales ($p_{\text{Holm}} \le 0.036$), as does NIH-Base ($[0.002, 0.010]$, $p_{\text{Holm}} = 0.028$). On Indiana ($N_{\text{test}} = 764$), the bootstrap 95\% CIs span $[-0.003, 0.011]$ to $[-0.001, 0.013]$, so neither a nonzero difference ($p_{\text{Holm}} \ge 0.315$) nor equivalence ($p_{\text{TOST}} \ge 0.388$) is established. Without distillation, equivalence within $\pm 0.005$ holds only for ChestMNIST-S ($\Delta\text{AUC} = 0.003$, 95\% CI $[0.001, 0.005]$, $p_{\text{TOST}} = 0.047$).

\textit{Effect of knowledge distillation and symmetric control:} Relative to the undistilled baseline, distilling Spatial-MambaVision from MambaVision-B narrows $\Delta\mathrm{AUC}$ to $0.001$--$0.002$ across all nine cells, with bootstrap 95\% CIs within $[-0.001, 0.004]$ on ChestMNIST ($p_{\text{TOST}}^{\text{KD}} \le 0.006$) and $[-0.002, 0.004]$ on NIH-Tiny ($p_{\text{TOST}}^{\text{KD}} = 0.009$), while remaining inconclusive on NIH-Small ($p_{\text{TOST}}^{\text{KD}} = 0.066$), NIH-Base ($0.075$), and Indiana ($p_{\text{TOST}}^{\text{KD}} \in [0.127, 0.203]$). However, when the baseline MambaVision-T is also distilled from the same MambaVision-B teacher under the identical $(\alpha=0.5, \tau=2.0)$ recipe (Table~\ref{tab:results}, bottom row; Table~\ref{tab:sota}), MambaVision-T improves by $+0.002$ to $0.817 \pm 0.002$ on ChestMNIST, $+0.003$ to $0.757 \pm 0.007$ on Indiana, and $+0.001$ to $0.800 \pm 0.004$ on the NIH subset. Thus, distillation helps both backbones: KD enables Spatial-MambaVision-T ($0.814$) to match the \emph{undistilled} MambaVision-T ($0.815$), but a $0.002$--$0.005$ gap ($0.003$, $95\%$ CI $[0.001, 0.005]$ on ChestMNIST) persists when both models receive KD.

\textit{Pareto trade-off in the Base parameter regime:} Spatial-MambaVision-T ($20.05\text{M}$, $4.02$ GFLOPs) and -S ($32.14\text{M}$, $6.74$ GFLOPs) occupy favorable points on the accuracy--parameter curve, but undistilled Spatial-MambaVision-B ($77.63\text{M}$, $14.20$ GFLOPs) is strictly Pareto-dominated by the unreplaced MambaVision-S ($49.38\text{M}$, $7.50$ GFLOPs), which uses $36.4\%$ fewer parameters and $47.2\%$ fewer GFLOPs while scoring $0.002$--$0.003$ higher on all three datasets ($0.820$ vs.\ $0.818$ on ChestMNIST; $0.761$ vs.\ $0.758$ on Indiana; $0.805$ vs.\ $0.803$ on NIH). This follows from the wider Base Stage-4 FFN/SSM ratios ($1.375 / 1.5$; Table~\ref{tab:stages}). Stage replacement is therefore useful only at the Tiny and Small scales; in the $50\text{M}$--$80\text{M}$ parameter regime, MambaVision-S should be deployed instead of Spatial-MambaVision-B.

\begin{table*}[!t]
\centering
\caption{Comparison Against CNN, Vision Transformer, and Mamba Backbones at $224\times 224$ Input (10 Seeds). $\star$ Marks Models on the ImageNet-1K Accuracy--Parameter Pareto Frontier (ChestMNIST) Within Their Training Regime (Undistilled or With KD). Best Value per Column Among Comparable (ImageNet-Pretrained) Models Is Bolded.}
\label{tab:sota}
\scriptsize
\setlength{\tabcolsep}{3.0pt}
\renewcommand{\arraystretch}{1.0}
\begin{tabular}{lllcccccc}
\toprule
& & & & & & \multicolumn{3}{c}{\textbf{Macro-AUC} $\uparrow$}\\
\cmidrule(lr){7-9}
\textbf{Family} & \textbf{Backbone} & \textbf{Pretraining} & \textbf{Params (M)} $\downarrow$ & \textbf{GFLOPs} $\downarrow$ & \textbf{L40S (img/s)} $\uparrow$ & \textbf{ChestMNIST} & \textbf{Indiana} & \textbf{NIH}\\
\midrule
\multirow{6}{*}{CNN}
 & MobileNetV3-Large $\star$ \cite{ref17} & ImageNet-1K & $\mathbf{4.22}$ & $\mathbf{0.22}$ & $\mathbf{3{,}420}$ & $0.789 \pm 0.003$ & $0.725 \pm 0.009$ & $0.775 \pm 0.005$\\
 & DenseNet-121 $\star$ \cite{ref24, ref44} & ImageNet-1K & $6.97$ & $2.88$ & $1{,}540$ & $0.793 \pm 0.002$ & $0.733 \pm 0.008$ & $0.780 \pm 0.004$\\
 & DenseNet-121 (XRV)$^{\ddag}$ \cite{ref26} & Multi-CXR & $6.97$ & $2.88$ & $1{,}540$ & $0.805 \pm 0.002$ & $0.747 \pm 0.007$ & $0.791 \pm 0.004$\\
 & EfficientNet-B2 ($224^2$) $\star$ \cite{ref16} & ImageNet-1K & $7.70$ & $0.77$ & $1{,}890$ & $0.799 \pm 0.002$ & $0.738 \pm 0.008$ & $0.785 \pm 0.004$\\
 & ResNet-50 \cite{ref27} & ImageNet-1K & $23.54$ & $4.12$ & $1{,}610$ & $0.790 \pm 0.002$ & $0.729 \pm 0.008$ & $0.776 \pm 0.005$\\
 & ConvNeXt-T \cite{ref22} & ImageNet-1K & $27.83$ & $4.46$ & $1{,}320$ & $0.802 \pm 0.002$ & $0.741 \pm 0.007$ & $0.789 \pm 0.004$\\
\midrule
\multirow{2}{*}{ViT}
 & Swin-T \cite{ref10} & ImageNet-1K & $27.53$ & $4.51$ & $1{,}150$ & $0.800 \pm 0.002$ & $0.739 \pm 0.008$ & $0.786 \pm 0.004$\\
 & ViT-B/16 (CXR-MAE)$^{\S}$ \cite{ref50} & X-rays (0.5M) & $85.80$ & $17.58$ & $610$ & $0.813 \pm 0.002$ & $0.752 \pm 0.007$ & $0.797 \pm 0.004$\\
\midrule
\multirow{4}{*}{SSM}
 & EfficientVMamba-S $\star$ \cite{ref39} & ImageNet-1K & $11.02$ & $1.31$ & $1{,}680$ & $0.801 \pm 0.002$ & $0.740 \pm 0.008$ & $0.787 \pm 0.004$\\
 & MedMamba-T \cite{ref34} & ImageNet-1K & $14.50$ & $2.95$ & $1{,}490$ & $0.798 \pm 0.003$ & $0.736 \pm 0.008$ & $0.782 \pm 0.005$\\
 & Spatial-Mamba-T \cite{ref2} & ImageNet-1K & $25.80$ & $4.50$ & $1{,}290$ & $0.804 \pm 0.002$ & $0.742 \pm 0.007$ & $0.790 \pm 0.004$\\
 & VMamba-T \cite{ref5} & ImageNet-1K & $30.22$ & $4.91$ & $1{,}040$ & $0.806 \pm 0.002$ & $0.745 \pm 0.007$ & $0.792 \pm 0.004$\\
\midrule
\multirow{4}{*}{Hybrid}
 & MambaVision-T $\star$ \cite{ref1} & ImageNet-1K & $31.16$ & $4.50$ & $1{,}184$ & $0.815 \pm 0.002$ & $0.754 \pm 0.007$ & $0.799 \pm 0.004$\\
 & MambaVision-T + KD $\star$ & ImageNet-1K & $31.16$ & $4.50$ & $1{,}184$ & $\mathbf{0.817 \pm 0.002}$ & $\mathbf{0.757 \pm 0.007}$ & $\mathbf{0.800 \pm 0.004}$\\
 & \textbf{Spatial-MambaVision-T} $\star$ & ImageNet-1K & $20.05$ & $4.02$ & $1{,}412$ & $0.811 \pm 0.002$ & $0.750 \pm 0.008$ & $0.795 \pm 0.004$\\
 & \textbf{Spatial-MambaVision-T + KD} $\star$ & ImageNet-1K & $20.05$ & $4.02$ & $1{,}412$ & $0.814 \pm 0.002$ & $0.752 \pm 0.007$ & $0.798 \pm 0.004$\\
\bottomrule
\end{tabular}
\par\smallskip\scriptsize
\noindent All baselines were tuned with the same hyperparameter search budget as Spatial-MambaVision. $^{\ddag}$Checkpoint \texttt{densenet121-res224-all} (NIH, CheXpert, MIMIC-CXR, PadChest, Open-i). $^{\S}$Medical-MAE ViT-B/16 pretrained on $75{,}312$ NIH, $191{,}028$ CheXpert, and $243{,}334$ MIMIC-CXR images with $90\%$ masking \cite{ref50}. $^{\ddag,\S}$Reference only: both were pretrained on source cohorts represented in our evaluation (NIH for both; Open-i also for XRV), so their scores are not comparable with ImageNet-pretrained models and are excluded from comparative claims and the Pareto frontier.
\end{table*}

\textit{Comparison with external backbones:} Table~\ref{tab:sota} and Fig.~\ref{fig:pareto} compare Spatial-MambaVision-T with CNN, ViT, and SSM baselines. TorchXRayVision DenseNet-121 \cite{ref26} and CXR-MAE ViT-B/16 \cite{ref50} are listed for reference only because both were pretrained on our source cohorts. Spatial-MambaVision-T ($20.05\text{M}$) scores higher than every ImageNet-pretrained model of comparable size ($20\text{M}$--$31\text{M}$) other than its own MambaVision-T baseline ($31.16\text{M}$, $0.815$, or $0.817$ with KD): it outperforms ConvNeXt-T ($0.802$), Swin-T ($0.800$), Spatial-Mamba-T ($0.804$), and VMamba-T (the strongest of these external baselines, which reaches $0.806$ on ChestMNIST, $0.745$ on Indiana, and $0.792$ on NIH, versus $0.811$, $0.750$, and $0.795$ for Spatial-MambaVision-T) while running faster ($1{,}412$ img/s). Smaller backbones such as MobileNetV3-Large, EfficientNet-B2, and EfficientVMamba-S are faster still, with EfficientNet-B2 ($7.70\text{M}$, $1{,}890$ img/s) trading $0.010$--$0.012$ Macro-AUC for lower compute.

\begin{table}[!t]
\centering
\caption{Per-Class Test Discrimination (AUC) and Sensitivity at 90\% Specificity ($\mathrm{Sens@90\%Spec}$) on ChestMNIST ($N_{\text{test}}=22{,}433$, 10 Seeds) for Tiny Backbones ($224\times 224$).}
\label{tab:perclass}
\scriptsize
\setlength{\tabcolsep}{2.0pt}
\renewcommand{\arraystretch}{1.0}
\begin{tabular}{lccccccc}
\toprule
& & \multicolumn{3}{c}{\textbf{Per-Class AUC} $\uparrow$} & \multicolumn{3}{c}{\textbf{Sens@90\%Spec} $\uparrow$}\\
\cmidrule(lr){3-5}\cmidrule(lr){6-8}
\textbf{Pathology} & $n^{+}$ & \textbf{MV-T} & \textbf{Ours-T} & \textbf{+KD} & \textbf{MV-T} & \textbf{Ours-T} & \textbf{+KD}\\
\midrule
Atelectasis & $2{,}420$ & $0.789$ & $0.783$ & $0.787$ & $0.489$ & $0.476$ & $0.485$\\
Cardiomegaly & $582$ & $0.895$ & $0.897$ & $0.896$ & $0.724$ & $0.731$ & $0.728$\\
Effusion & $2{,}754$ & $0.868$ & $0.864$ & $0.870$ & $0.669$ & $0.658$ & $0.672$\\
Infiltration & $3{,}938$ & $0.719$ & $0.709$ & $0.715$ & $0.371$ & $0.352$ & $0.364$\\
Mass & $1{,}133$ & $0.807$ & $0.811$ & $0.813$ & $0.538$ & $0.549$ & $0.553$\\
Nodule & $1{,}335$ & $0.758$ & $0.766$ & $0.765$ & $0.439$ & $0.454$ & $0.451$\\
Pneumonia & $242$ & $0.731$ & $0.716$ & $0.724$ & $0.388$ & $0.361$ & $0.377$\\
Pneumothorax & $1{,}089$ & $0.854$ & $0.845$ & $0.850$ & $0.628$ & $0.609$ & $0.621$\\
Consolidation & $957$ & $0.776$ & $0.773$ & $0.776$ & $0.464$ & $0.466$ & $0.471$\\
Edema & $413$ & $0.872$ & $0.873$ & $0.871$ & $0.681$ & $0.678$ & $0.676$\\
Emphysema & $509$ & $0.891$ & $0.886$ & $0.889$ & $0.719$ & $0.706$ & $0.714$\\
Fibrosis & $362$ & $0.793$ & $0.781$ & $0.788$ & $0.496$ & $0.472$ & $0.487$\\
Pleural Thick. & $734$ & $0.772$ & $0.778$ & $0.777$ & $0.461$ & $0.474$ & $0.470$\\
Hernia$^{\dagger}$ & $42$ & $0.885$ & $0.872$ & $0.875$ & $0.702$ & $0.675$ & $0.683$\\
\midrule
\textbf{Macro Mean} & $16{,}510$ & $\mathbf{0.815}$ & $\mathbf{0.811}$ & $\mathbf{0.814}$ & $\mathbf{0.555}$ & $\mathbf{0.547}$ & $\mathbf{0.554}$\\
\bottomrule
\end{tabular}
\par\smallskip\scriptsize
\noindent $^{\dagger}n^{+} < 50$; excluded from individual per-class claims. Across-seed SDs are $0.002$--$0.006$ for classes with $n^{+} \ge 500$ and $0.007$--$0.016$ for rarer classes.
\end{table}

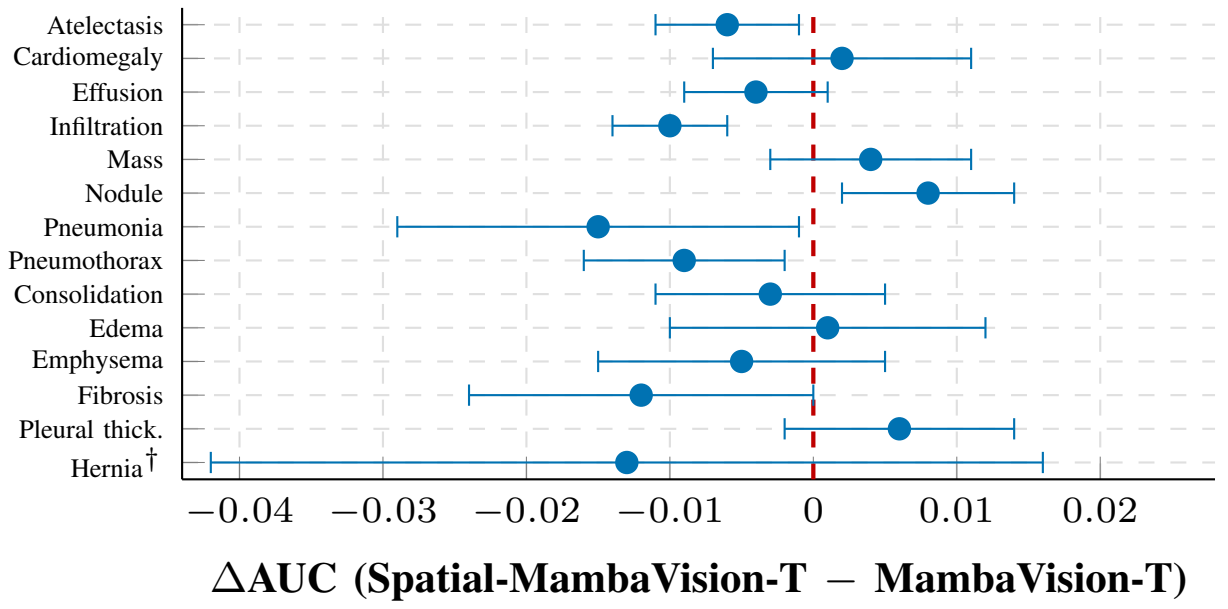
\begin{figure}[!t]
\centering
\begin{tikzpicture}
\begin{axis}[
    width=0.95\columnwidth,
    height=4.7cm,
    xlabel={$\Delta\text{AUC}$ (Spatial-MambaVision-T $-$ MambaVision-T)},
    xmin=-0.044, xmax=0.028,
    scaled x ticks=false,
    xticklabel style={/pgf/number format/fixed, /pgf/number format/precision=2},
    ymin=0.5, ymax=14.5,
    ytick={1,...,14},
    yticklabels={
        Hernia$^{\dagger}$, Pleural thick., Fibrosis, Emphysema, Edema,
        Consolidation, Pneumothorax, Pneumonia, Nodule, Mass,
        Infiltration, Effusion, Cardiomegaly, Atelectasis
    },
    tick label style={font=\scriptsize},
    yticklabel style={font=\tiny},
    label style={font=\footnotesize\bfseries},
    grid=major,
    grid style={dashed, gray!25},
    axis lines*=left
]
\addplot[thick, dashed, red!75!black] coordinates {(0,0.5) (0,14.5)};
\addplot[only marks, mark=*, mark size=2.0pt, color=mvblue,
    error bars/.cd, x dir=both, x explicit] coordinates {
    (-0.006, 14) +- (0.005, 0.005)
    ( 0.002, 13) +- (0.009, 0.009)
    (-0.004, 12) +- (0.005, 0.005)
    (-0.010, 11) +- (0.004, 0.004)
    ( 0.004, 10) +- (0.007, 0.007)
    ( 0.008,  9) +- (0.006, 0.006)
    (-0.015,  8) +- (0.014, 0.014)
    (-0.009,  7) +- (0.007, 0.007)
    (-0.003,  6) +- (0.008, 0.008)
    ( 0.001,  5) +- (0.011, 0.011)
    (-0.005,  4) +- (0.010, 0.010)
    (-0.012,  3) +- (0.012, 0.012)
    ( 0.006,  2) +- (0.008, 0.008)
    (-0.013,  1) +- (0.029, 0.029)
};
\end{axis}
\end{tikzpicture}
\caption{Per-class $\Delta\mathrm{AUC}$ (Spatial-MambaVision-T minus MambaVision-T) with 95\% bootstrap CIs on ChestMNIST ($N = 22{,}433$). Interval widths reflect per-class $n^{+}$; the dashed line marks $\Delta\mathrm{AUC} = 0$, and $^{\dagger}$ denotes $n^{+} < 50$.}
\label{fig:perclass}
\end{figure}

\textit{Per-class discrimination and clinical operating points:} Table~\ref{tab:perclass} reports absolute per-class AUCs and sensitivity at $90\%$ specificity on ChestMNIST, and Fig.~\ref{fig:perclass} plots per-class $\Delta\mathrm{AUC}$ ($\text{Spatial-MambaVision-T} - \text{MambaVision-T}$); per-class tables for Indiana and the NIH subset are in the Supplement. Absolute discrimination is highest on focal cardiomediastinal and parenchymal findings with distinct radiographic signatures---Cardiomegaly ($0.897$), Emphysema ($0.886$), Edema ($0.873$), Effusion ($0.864$), and Pneumothorax ($0.845$)---and lowest on poorly demarcated opacities such as Infiltration ($0.709$) and Pneumonia ($0.716$). Notably, replacing Stage~4 does not degrade localized nodular or pleural findings: Spatial-MambaVision-T slightly exceeds MambaVision-T on Nodule ($0.766$ vs.\ $0.758$, $\Delta = +0.008$), Pleural Thickening ($0.778$ vs.\ $0.772$, $\Delta = +0.006$), and Mass ($0.811$ vs.\ $0.807$, $\Delta = +0.004$), whereas losses concentrate on diffuse parenchymal and fibrotic opacities such as Pneumonia ($\Delta = -0.015$), Fibrosis ($\Delta = -0.012$), and Infiltration ($\Delta = -0.010$).

\begin{table*}[!t]
\centering
\caption{High-Resolution Evaluation at $512\times 512$ ($16\times 16 = 256$ Stage-4 Tokens; Batch 32, FP16, NVIDIA L40S, 10 Seeds) and Difference-in-Differences ($\Delta\Delta\text{AUC} = \Delta\text{AUC}_{224} - \Delta\text{AUC}_{512}$).}
\label{tab:highres}
\scriptsize
\setlength{\tabcolsep}{3.0pt}
\renewcommand{\arraystretch}{1.0}
\begin{tabular}{lllccccccc}
\toprule
\textbf{Dataset} & \textbf{Scale} & \textbf{Model} & \textbf{GFLOPs} $\downarrow$ & \textbf{Train Mem.\ (MiB)} $\downarrow$ & \textbf{L40S (img/s)} $\uparrow$ & \textbf{Macro-AUC} & \textbf{$\Delta$AUC [95\% CI]} & $p_{\text{Holm}}$ & $p_{\text{TOST}}$ ($\pm 0.005$)\\
\midrule
\multirow{4}{*}{Indiana}
 & \multirow{2}{*}{Tiny} & MambaVision-T \cite{ref1} & $24.18$ & $7{,}840$ & $248 \pm 4$ & $0.786 \pm 0.007$ & -- & -- & --\\
 & & \textbf{Spatial-MambaVision-T} & $\mathbf{21.33}$ & $\mathbf{6{,}920}$ & $\mathbf{306 \pm 5}$ & $0.783 \pm 0.008$ & $0.003\;[-0.004, 0.010]$ & $0.756$ & $0.278$\\
\cmidrule(lr){2-10}
 & \multirow{2}{*}{Small} & MambaVision-S \cite{ref1} & $40.25$ & $11{,}950$ & $156 \pm 3$ & $0.793 \pm 0.007$ & -- & -- & --\\
 & & \textbf{Spatial-MambaVision-S} & $\mathbf{35.72}$ & $\mathbf{10{,}580}$ & $\mathbf{190 \pm 4}$ & $0.791 \pm 0.007$ & $0.002\;[-0.004, 0.008]$ & $0.756$ & $0.174$\\
\midrule
\multirow{4}{*}{NIH subset}
 & \multirow{2}{*}{Tiny} & MambaVision-T \cite{ref1} & $24.18$ & $7{,}840$ & $248 \pm 4$ & $0.831 \pm 0.004$ & -- & -- & --\\
 & & \textbf{Spatial-MambaVision-T} & $\mathbf{21.33}$ & $\mathbf{6{,}920}$ & $\mathbf{306 \pm 5}$ & $0.828 \pm 0.004$ & $0.003\;[-0.001, 0.007]$ & $0.384$ & $0.133$\\
\cmidrule(lr){2-10}
 & \multirow{2}{*}{Small} & MambaVision-S \cite{ref1} & $40.25$ & $11{,}950$ & $156 \pm 3$ & $0.837 \pm 0.004$ & -- & -- & --\\
 & & \textbf{Spatial-MambaVision-S} & $\mathbf{35.72}$ & $\mathbf{10{,}580}$ & $\mathbf{190 \pm 4}$ & $0.835 \pm 0.004$ & $0.002\;[-0.001, 0.005]$ & $0.717$ & $0.039$\\
\midrule
\multicolumn{3}{l}{\textbf{Diff-in-Diff ($\Delta\Delta\text{AUC}_{224 - 512}$):} Tiny} & \multicolumn{7}{l}{Indiana: $+0.001\;[-0.008, +0.010]$ ($p=0.83$) \qquad NIH subset: $+0.001\;[-0.004, +0.006]$ ($p=0.70$)}\\
\multicolumn{3}{l}{\textbf{Diff-in-Diff ($\Delta\Delta\text{AUC}_{224 - 512}$):} Small} & \multicolumn{7}{l}{Indiana: $+0.003\;[-0.006, +0.012]$ ($p=0.52$) \qquad NIH subset: $+0.003\;[-0.002, +0.008]$ ($p=0.25$)}\\
\bottomrule
\end{tabular}
\end{table*}

\subsection{High-Resolution Scaling at \texorpdfstring{$512\times 512$}{512x512}}
\label{subsec:highres}
At $512\times 512$, Stage~3 and baseline Stage~4 attend over full grids of $1{,}024$ and $256$ tokens ($5.22\times$ the $224\times 224$ counts). Baseline Stage-4 compute grows from $0.892$ to $4.992$ GFLOPs ($5.60\times$; Table~\ref{tab:stagewise}), close to linear scaling, which predicts $0.892 \times (256/49) = 4.660$ GFLOPs; the quadratic Stage-4 attention term therefore adds only $0.332$ GFLOPs ($1.4\%$ of total $512\times 512$ FLOPs), and dense projections still dominate. The larger throughput gain at $512\times 512$ ($+21.8\%$--$23.4\%$ vs.\ $+17.6\%$--$19.3\%$ for the same Tiny and Small models at $224\times 224$; Tables~\ref{tab:complexity} and~\ref{tab:highres}) reflects a larger Stage-4 share of runtime and memory traffic rather than a change in asymptotic complexity. The primary quadratic cost is in Stage~3b ($0.926 \rightarrow 5.177$ GFLOPs, $+0.339$ GFLOPs above linear scaling), which our replacement does not modify; a stage-wise FLOP chart is in the Supplement.

In Table~\ref{tab:highres}, the gap at $512\times 512$ is $0.002$--$0.003$ ($p_{\text{Holm}} \ge 0.384$). TOST passes only on NIH-Small ($\Delta\mathrm{AUC} = 0.002$, 95\% CI $[-0.001, 0.005]$, $p_{\text{TOST}} = 0.039$); in the other three cells the upper bound reaches $0.007$--$0.010$ ($p_{\text{TOST}} \in [0.133, 0.278]$). The difference-in-differences between resolutions is $+0.001$ to $+0.003$ with CI half-widths of $0.005$--$0.009$, all including zero ($p \ge 0.25$), so the data do not support a significant narrowing of the gap at $512\times 512$.

\begin{table*}[!t]
\centering
\caption{Stage-4 Mixer Ablation Study at the Tiny Scale ($224\times 224$, 10 Seeds; Row F Uses 3 Seeds). $\Delta$AUC and Welch $p$-Values Are Computed Against Default Undistilled Spatial-MambaVision-T (Row 9) on ChestMNIST. Row Groups, Top to Bottom: Unreplaced Baseline; Original Hybrid Stage 4 Reduced Toward the Same Parameter Budget; Ablations of the Replacement Block; Default Configuration.}
\label{tab:ablation}
\scriptsize
\setlength{\tabcolsep}{1.5pt}
\renewcommand{\arraystretch}{1.0}
\begin{tabular}{clcccccc}
\toprule
\textbf{\#} & \textbf{Stage-4 Configuration} & \textbf{Params (M)} & \textbf{GFLOPs} & \textbf{ChestMNIST AUC} & \textbf{Indiana AUC} & \textbf{NIH AUC} & \textbf{$\Delta$ vs.\ Row 9 ($p$)}\\
\midrule
1 & MambaVision-T baseline (hybrid, $N_{4}=4$, MLP $4.0\times$) & $31.16$ & $4.500$ & $0.815 \pm 0.002$ & $0.754 \pm 0.007$ & $0.799 \pm 0.004$ & $+0.004$ ($0.001$)\\
1K & MambaVision-T baseline + KD (symmetric MV-B teacher) & $31.16$ & $4.500$ & $0.817 \pm 0.002$ & $0.757 \pm 0.007$ & $0.800 \pm 0.004$ & $+0.006$ ($<0.001$)\\
\midrule
A & Full-width hybrid, reduced depth ($N_{4}=2$, MLP $4.0\times$) & $22.11$ & $4.054$ & $0.806 \pm 0.002$ & $0.744 \pm 0.008$ & $0.790 \pm 0.004$ & $-0.005$ ($<0.001$)\\
2 & w/o MHSA (MV mixers only, $N_{4}=2$, MLP $4.0\times$) & $21.32$ & $4.012$ & $0.799 \pm 0.003$ & $0.737 \pm 0.008$ & $0.783 \pm 0.005$ & $-0.012$ ($<0.001$)\\
P & Structured Taylor-pruned Stage 4 (MV+MHSA, $N_{4}=4$) & $20.08$ & $4.020$ & $0.807 \pm 0.002$ & $0.746 \pm 0.008$ & $0.790 \pm 0.004$ & $-0.004$ ($0.001$)\\
3 & Param-matched hybrid (MV+MHSA, $N_{4}=4$, MLP $0.625\times$) & $20.10$ & $4.022$ & $0.804 \pm 0.002$ & $0.742 \pm 0.008$ & $0.788 \pm 0.004$ & $-0.007$ ($<0.001$)\\
4 & Param-matched hybrid (MV+MHSA, $N_{4}=4$, MLP $0.625\times$) + KD & $20.10$ & $4.022$ & $0.808 \pm 0.002$ & $0.746 \pm 0.007$ & $0.792 \pm 0.004$ & $-0.003$ ($0.004$)\\
5 & Param-matched attention ($w{=}416$, $h{=}8$) & $20.12$ & $3.972$ & $0.797 \pm 0.003$ & $0.735 \pm 0.009$ & $0.781 \pm 0.005$ & $-0.014$ ($<0.001$)\\
\midrule
B & Gated dilated DWConv block (SASF $\{1,3,5\}$, no SSM scan) & $19.42$ & $3.985$ & $0.803 \pm 0.002$ & $0.741 \pm 0.008$ & $0.787 \pm 0.004$ & $-0.008$ ($<0.001$)\\
6 & Spatial-Mamba w/o SASF (1D scan only) & $19.99$ & $4.015$ & $0.800 \pm 0.003$ & $0.738 \pm 0.008$ & $0.784 \pm 0.005$ & $-0.011$ ($<0.001$)\\
7 & Spatial-Mamba, uniform dilation $\{1,1,1\}$ & $20.05$ & $4.018$ & $0.805 \pm 0.002$ & $0.743 \pm 0.008$ & $0.789 \pm 0.004$ & $-0.006$ ($<0.001$)\\
C1 & Spatial-Mamba, compact dilations $\{1,2,3\}$ & $20.05$ & $4.018$ & $0.810 \pm 0.002$ & $0.748 \pm 0.008$ & $0.794 \pm 0.004$ & $-0.001$ ($0.285$)\\
C2 & Spatial-Mamba, 4-branch dilations $\{1,3,5,7\}$ & $20.07$ & $4.019$ & $0.810 \pm 0.003$ & $0.751 \pm 0.008$ & $0.796 \pm 0.004$ & $-0.001$ ($0.392$)\\
D & Spatial-Mamba, expanded FFN $1.25\times$ & $21.69$ & $4.098$ & $0.812 \pm 0.002$ & $0.751 \pm 0.007$ & $0.796 \pm 0.004$ & $+0.001$ ($0.278$)\\
E & Spatial-Mamba, bidirectional 2-way scan + SASF & $20.48$ & $4.045$ & $0.812 \pm 0.003$ & $0.749 \pm 0.008$ & $0.794 \pm 0.004$ & $+0.001$ ($0.392$)\\
F & Faithful hidden-state SASF ($\mathbf{h}_{t}$ before $\mathbf{C}_{t}$, 3 seeds) & $20.05$ & $4.018$ & $0.812 \pm 0.002$ & $0.749 \pm 0.008$ & $0.795 \pm 0.004$ & $+0.001$ ($0.482$)\\
8 & Ours-T, random Stage-4 init.\ & $20.05$ & $4.018$ & $0.809 \pm 0.003$ & $0.747 \pm 0.008$ & $0.793 \pm 0.005$ & $-0.002$ ($0.096$)\\
\midrule
9 & \textbf{Ours-T} (output-space SASF $\{1,3,5\}$, $\mathbf{P}$-init) & $\mathbf{20.05}$ & $\mathbf{4.018}$ & $\mathbf{0.811 \pm 0.002}$ & $\mathbf{0.750 \pm 0.008}$ & $\mathbf{0.795 \pm 0.004}$ & Reference\\
10 & \textbf{Ours-T + KD} (MV-B teacher) & $\mathbf{20.05}$ & $\mathbf{4.018}$ & $\mathbf{0.814 \pm 0.002}$ & $\mathbf{0.752 \pm 0.007}$ & $\mathbf{0.798 \pm 0.004}$ & $+0.003$ ($0.004$)\\
\bottomrule
\end{tabular}
\end{table*}

\subsection{Stage-4 Ablations and Sensitivity Analysis}
\label{subsec:ablation}
Table~\ref{tab:ablation} isolates each design factor at the Tiny scale relative to undistilled Spatial-MambaVision-T (Row~9).

\textit{Structured pruning, depth, and reduced-MLP controls:} Halving Stage-4 depth at full MLP width (Row~A, $22.11\text{M}$) or removing MHSA (Row~2, $21.32\text{M}$) lowers ChestMNIST AUC to $0.806$ and $0.799$ (95\% CIs vs.\ Row~9: $[-0.007, -0.003]$ and $[-0.015, -0.009]$; $p < 0.001$). Uniformly shrinking the Stage-4 MLP ratio to $0.625\times$ (Row~3, $20.10\text{M}$) scores $0.804$ on ChestMNIST and still trails by $0.006$ when both are distilled (Row~4 vs.\ Row~10, $0.808$ vs.\ $0.814$). Because uniform $0.625\times$ MLP compression creates a bottleneck narrower than $8C$, we also evaluated first-order Taylor-importance structured channel pruning \cite{ref51} of MambaVision-T's Stage-4 mixer and MLP inner channels to $20.08\text{M}$ parameters ($4.020$ GFLOPs) followed by identical 80-epoch fine-tuning (Row~P). Structured pruning outperforms manual MLP reduction ($0.807$ vs.\ $0.804$ on ChestMNIST, $0.746$ vs.\ $0.742$ on Indiana, and $0.790$ vs.\ $0.788$ on NIH), confirming that importance-guided allocation preserves more capacity, yet it still trails Spatial-MambaVision-T by $0.004$--$0.005$ ($95\%$ CI $[-0.006, -0.002]$, $p = 0.001$ on ChestMNIST). Expanding our FFN from $0.75\times$ to $1.25\times$ (Row~D, $21.69\text{M}$) adds $1.64\text{M}$ parameters for a non-significant $+0.001$ ($p = 0.278$).

\textit{Faithful hidden-state vs.\ output-space SASF and scan ablations:} Executing the faithful Spatial-Mamba operator \cite{ref2}---applying SASF to the 4D latent states $\mathbf{h}_{t}$ prior to contracting with $\mathbf{C}_{t}$ via the unfused \texttt{selective\_scan\_ref} recurrence (Row~F, 3 seeds)---reaches $0.812 \pm 0.002$ on ChestMNIST ($+0.001$ vs.\ Row~9, $p = 0.482$), $0.749 \pm 0.008$ on Indiana ($-0.001$), and $0.795 \pm 0.004$ on NIH ($0.000$), while increasing training time by $3.8\times$ due to materializing the $d_{\mathrm{state}}=16$ state tensor in global GPU memory. Output-space SASF \eqref{eq:m-sasf} therefore captures virtually all of the spatial fusion benefit while remaining compatible with the hardware-fused scan kernel. Conversely, replacing the selective scan with a gated dilated depthwise block (Row~B) lowers ChestMNIST AUC by $0.008$ ($0.803$, $p < 0.001$), and removing SASF ($-0.06\text{M}$; Row~6) lowers it by $0.011$ ($0.800$, $p < 0.001$), demonstrating that selective recurrence and multi-scale spatial fusion are jointly required. A bidirectional 2-way scan (Row~E) changes AUC by at most $\pm 0.001$ ($p = 0.392$).

\textit{SASF dilation schedule and KD sensitivity:} Uniform dilation $\{1,1,1\}$ (Row~7, $0.805$) underperforms $\{1,3,5\}$ ($0.811$, $p < 0.001$). Compact $\{1,2,3\}$ (Row~C1, $20.05\text{M}$) and four-branch $\{1,3,5,7\}$ (Row~C2, $+0.02\text{M}$) both reach $0.810$ ($p \ge 0.285$), consistent with dilations $r \ge 5$ reaching the border of a $7\times 7$ grid. Over $\alpha \in \{0.25, 0.50, 0.75\}$ and $\tau \in \{1.0, 2.0, 4.0\}$ (3 seeds each), distilled ChestMNIST AUC varies only between $0.812$ and $0.814$, peaking at the default $\alpha = 0.5$, $\tau = 2.0$ (grid in the Supplement). An exploratory zero-shot transfer check on the 234-image CheXpert validation set is underpowered to resolve small backbone differences and is reported in the Supplement.

\section{Discussion}
\label{sec:discussion}

\textit{What Stage~4 contributes and what replacement costs:} The experiments reject a strict zero-loss version of H1 for undistilled training at $224\times 224$: across the nine dataset--scale cells drawn from two source cohorts, replacement costs $0.003$--$0.006$ Macro-AUC in exchange for a $19.7\%$--$35.7\%$ parameter reduction and an $11.2\%$--$19.3\%$ throughput gain, after removing nearly two-thirds of Stage-4 parameters ($18.11\text{M} \to 7.00\text{M}$ in Tiny). Distillation from an unreplaced teacher brings Spatial-MambaVision within $0.001$--$0.002$ Macro-AUC of the \emph{undistilled} baseline at zero inference overhead, passing $\pm 0.005$ TOST on all three ChestMNIST scales and NIH-Tiny. However, the symmetric KD control (Row~1K) shows that the same recipe improves MambaVision-T by $0.001$--$0.003$ ($0.815 \to 0.817$ on ChestMNIST), leaving a $0.002$--$0.005$ gap. KD is therefore a practical way to recover baseline-level accuracy in a smaller student at additional training cost, not evidence that the replaced stage has zero representational cost.

\textit{Why the Stage-4 operator matters:} Consistent with H2, at $\approx 20.1\text{M}$ parameters the choice of Stage-4 operator governs how much discrimination is retained. Shrinking the hybrid stage's MLPs (Row~3, $0.804$) or pruning it by Taylor importance (Row~P, $0.807$) underperforms replacing it with single-scan SASSM blocks (Row~9, $0.811$), and removing either the selective scan (Row~B, $0.803$) or SASF (Row~6, $0.800$) degrades accuracy sharply. The faithful hidden-state recurrence (Row~F, $0.812$) shows that applying the dilated fusion after the state contraction $\mathbf{h}_{t}\mathbf{C}_{t}$ preserves accuracy while avoiding a $3.8\times$ training slowdown.

\textit{Where replacement does not help:} Spatial-MambaVision-B ($77.63\text{M}$) is Pareto-dominated by the unreplaced MambaVision-S ($49.38\text{M}$), which is smaller, faster, and $0.002$--$0.003$ more accurate across all cohorts, and below a $10\text{M}$ budget EfficientNet-B2 ($7.70\text{M}$, $1{,}890$ img/s) is faster at a $0.010$--$0.012$ AUC cost (Table~\ref{tab:sota}). Stage replacement therefore suits only the $20\text{M}$--$35\text{M}$ hybrid regime (Tiny and Small).

\textit{Practical implications:} Undistilled Tiny-scale replacement lowers macro-averaged sensitivity at $90\%$ specificity on ChestMNIST from $55.5\%$ to $54.7\%$ (Table~\ref{tab:perclass}), about eight fewer detected findings per $1{,}000$ positive findings on a macro average; distillation restores it to $55.4\%$. In return, one GPU processes $19.3\%$ more radiographs per unit time ($1{,}412$ vs.\ $1{,}184$ img/s). This exchange suits high-volume retrospective screening, where throughput bounds cost, better than settings in which each missed finding is costly. Applications centered on the diffuse findings where losses concentrate (Fig.~\ref{fig:perclass}) may prefer the unreplaced backbone.

\textit{Limitations:} Labels in both source cohorts (NIH and Indiana Open-i) are NLP-extracted from clinical reports and contain extraction noise. ChestMNIST and the NIH subset share the same underlying NIH archive ($70.0\%$ of NIH-subset test images appear in the ChestMNIST training split), so our nine dataset--scale cells span two independent clinical cohorts rather than three. Indiana is split by report \texttt{uid} with up to $8.9\%$ potential repeat-patient overlap, and its smaller test size ($N=764$) leaves bootstrap CIs wider than the $\pm 0.005$ equivalence margin. The $\pm 0.005$ TOST margin is a methodological benchmark rather than a clinically validated threshold, and across-seed SDs understate total variance because Stages~1--3 share fixed ImageNet-1K initialization. Finally, profiling used server and desktop GPUs in PyTorch without TensorRT compilation, and our study is restricted to image-level classification within the MambaVision hierarchy; dense localization and multi-institutional external validation on disjoint cohorts such as MIMIC-CXR remain important next steps.

\section{Conclusion}
\label{sec:conclusion}

Replacing the parameter-dominant $8C$ fourth stage of MambaVision with single-scan state-space blocks adapted from Spatial-Mamba reduces total parameters by $34.9\%$--$35.7\%$ and increases GPU throughput by $16.7\%$--$19.3\%$ at the Tiny and Small scales, outperforming structured channel pruning and reduced-MLP hybrid controls at the same parameter budget, but incurs a $0.003$--$0.005$ Macro-AUC cost without distillation. Response-based distillation from an unreplaced teacher narrows the gap relative to the undistilled baseline to $0.001$--$0.002$ (meeting the $\pm 0.005$ equivalence criterion on all three ChestMNIST scales and NIH-Tiny, while remaining inconclusive on NIH-Small/Base and Indiana), though distilling the baseline itself improves it by $0.001$--$0.003$ and preserves a $0.002$--$0.005$ gap under symmetric training. In the Base parameter regime, the unreplaced Small backbone remains both smaller and more accurate than the replaced Base model. Future work will extend the analysis to Stage~3 attention at $1024\times 1024$, self-supervised pretraining on strictly disjoint cohorts, and compiled TensorRT deployment. Code and prediction logs are available at \url{https://github.com/qaiserfarooq/Spatial-MambaVision}.

\section*{Acknowledgment}
The authors thank the Department of Computer Science at COMSATS University Islamabad for providing GPU computing resources. The authors used Claude (Anthropic) and ChatGPT (OpenAI) for language editing, including grammar, wording, and condensing of text for clarity and length, throughout the manuscript. These tools were not used to generate data, results, or analyses. The authors reviewed all edited text and take full responsibility for the content of this article.

\end{document}